\pdfoutput=1

\documentclass[11pt]{article}

\usepackage[final]{acl}

\usepackage{times}
\usepackage{latexsym}

\usepackage[T1]{fontenc}

\usepackage[utf8]{inputenc}

\usepackage{microtype}

\usepackage{inconsolata}

\usepackage{graphicx}
\usepackage{inconsolata}
\usepackage{xspace, soul}
\usepackage{threeparttable}
\usepackage{booktabs}
\usepackage{multirow}
\usepackage{amsmath}
\usepackage{pifont}
\usepackage{amssymb,enumitem}
\usepackage{comment}
\usepackage{hyperref}
\usepackage{subfig}
\usepackage{caption}
\usepackage{wrapfig}
\usepackage{listings}
\usepackage{adjustbox}
\usepackage{marvosym}
\newcommand{\cmark}{\ding{51}}
\newcommand{\xmark}{\ding{55}}  
\title{\dataset: Strategic Utilization of Visuals for Robust Multimodal E-commerce Models}

\author{
 \textbf{Xinyi Ling\textsuperscript{1}},
 \textbf{Hanwen Du\textsuperscript{1}},
 \textbf{Zhihui Zhu\textsuperscript{1}},
 \textbf{Xia Ning\textsuperscript{1 2 3 \Letter}}
\\
 \textsuperscript{1}Department of Computer Science and Engineering, The Ohio State University\\
 \textsuperscript{2}Translational Data Analytics Institute, The Ohio State University\\
 \textsuperscript{3}Department of Biomedical Informatics, The Ohio State University
\\
   \texttt{\{ling.303, du.1128, zhu.3440, ning.104\}@osu.edu}
 }

\newcommand{\PRP}{\mbox{$\mathop{\mathtt{PRP}}\limits$}\xspace}

\newcommand{\CP}{\mbox{$\mathop{\mathtt{CP}}\limits$}\xspace}

\newcommand{\SA}{\mbox{$\mathop{\mathtt{SA}}\limits$}\xspace}

\newcommand{\SR}{\mbox{$\mathop{\mathtt{SR}}\limits$}\xspace}

\newcommand{\MPC}{\mbox{$\mathop{\mathtt{MPC}}\limits$}\xspace}

\newcommand{\PSI}{\mbox{$\mathop{\mathtt{PSI}}\limits$}\xspace}

\newcommand{\AP}{\mbox{$\mathop{\mathtt{AP}}\limits$}\xspace}

\newcommand{\QA}{\mbox{$\mathop{\mathtt{BQA}}\limits$}\xspace}

\newcommand{\dataset}{\mbox{$\mathop{\mathtt{EcomMMMU}}\limits$}\xspace}
\newcommand{\method}{\mbox{$\mathop{\mathtt{SUMEI}}\limits$}\xspace}

\newcommand{\methodL}{\mbox{$\mathop{\mathtt{\method_{\text{Llava}}}}\limits$}\xspace}
\newcommand{\methodC}{\mbox{$\mathop{\mathtt{\method_{\text{CASLIE}}}}\limits$}\xspace}

\newcommand{\VSS}{\mbox{$\mathop{\mathtt{VSS}}\limits$}\xspace}
\newcommand{\GTS}{\mbox{$\mathop{\mathtt{GTS}}\limits$}\xspace}

\newcommand{\hard}{\mbox{\VSS}\xspace}

\newcommand{\VUA}{\mbox{$\mathop{\mathtt{\method{\text{-}}vua}}\limits$}\xspace}
\newcommand{\VUP}{\mbox{$\mathop{\mathtt{\method{\text{-}}vup}}\limits$}\xspace}
\newcommand{\VSM}{\mbox{$\mathop{\mathtt{\method{\text{-}}vsm}}\limits$}\xspace}

\lstdefinelanguage{json}{
    basicstyle=\ttfamily\small,
    numbers=none,
    stepnumber=1,
    numbersep=5pt,
    showstringspaces=false,
    breaklines=true,
    frame=single,
    backgroundcolor=\color{gray!3},
}

\begin{document}
\maketitle


\begin{abstract}
E-commerce platforms are rich in multimodal data, featuring a variety of images that depict product details.
However, this raises an important question: do these images always enhance product understanding, 
or can they sometimes introduce redundancy or degrade performance?
%
Existing datasets are limited in both scale and design, making it difficult to systematically examine this question.
To this end, we introduce \dataset, an e-commerce multimodal multitask understanding dataset with 406,190 samples and 8,989,510 images. 
\dataset is comprised of multi-image visual-language data designed with 8 essential tasks and 
a specialized \hard subset to benchmark 
the capability of multimodal large language models (MLLMs) to effectively utilize visual content.
%
Analysis on \dataset reveals that product images do not consistently improve performance and can, in some cases, degrade it. 
This indicates that MLLMs may struggle to effectively leverage rich visual content for e-commerce tasks.
Building on these insights, we propose \method, a data-driven method that strategically utilizes multiple images via predicting visual utilities before using them for downstream tasks.
Comprehensive experiments demonstrate the effectiveness and robustness of \method.
The data and code are available through \url{https://github.com/ninglab/EcomMMMU}.
\end{abstract}

\begin{figure*}
    \centering
    \subfloat[Sample from \dataset \label{fig:dataset_cases}]{\includegraphics[trim=0 0 10 0, clip, width=0.45\textwidth]{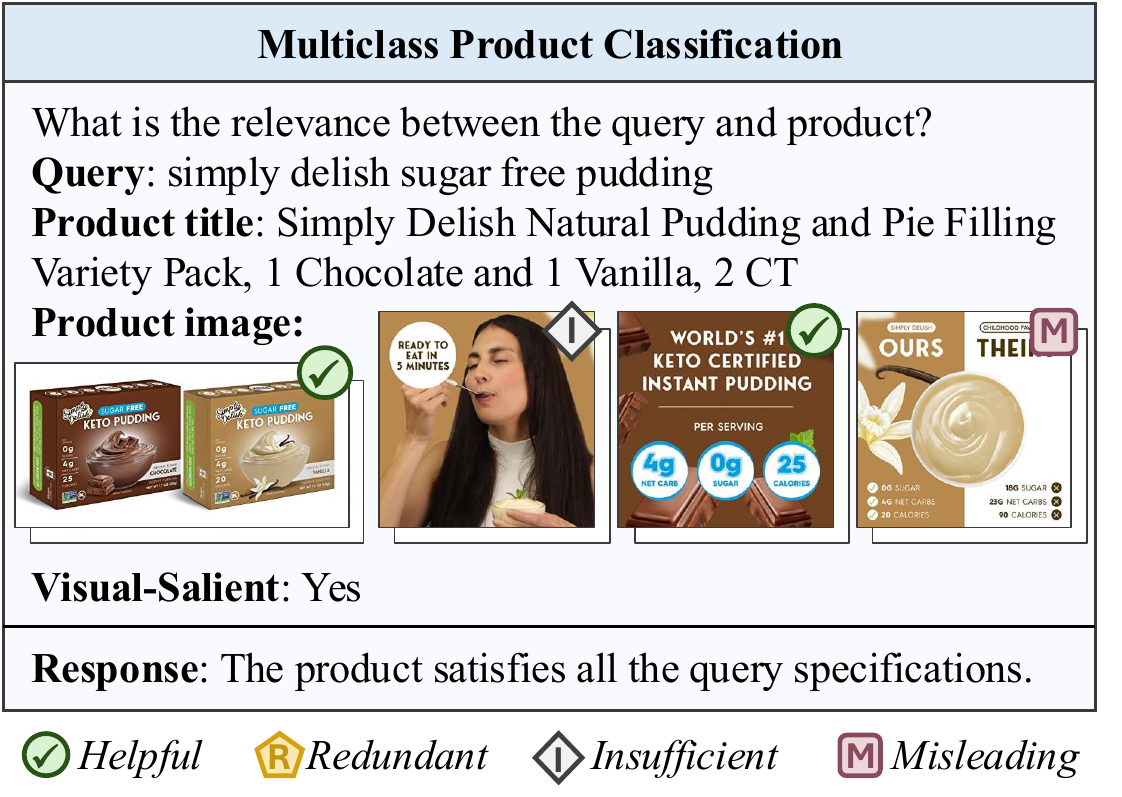}}
    \hspace{15pt}
    \subfloat[Workflow of Visual-salient Set Identification \label{fig:hard_identification}]
    {\includegraphics[width=0.45\textwidth]{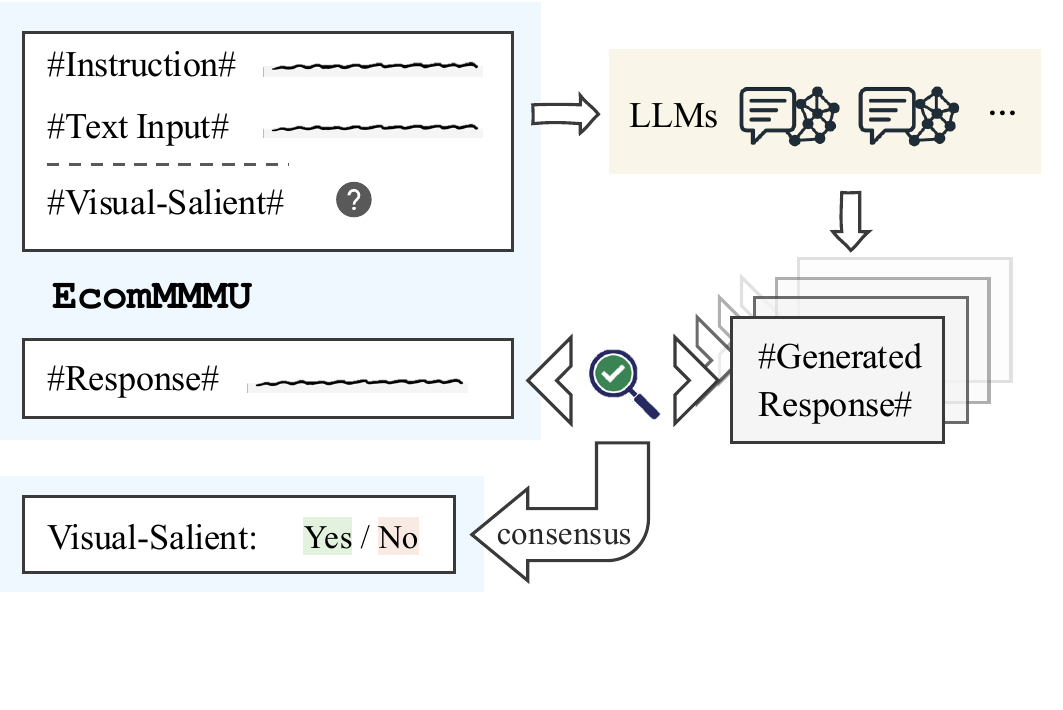}}
    \caption{\dataset overview. Visual-salient labels are pre-identified in \dataset.
    }
    \label{fig:dataset_overview}
\end{figure*}


%

\section{Introduction}
\label{sec:intro}

E-commerce platforms have become pivotal centers of consumer activities~\citep{kang2018sasrec, ni2019justifying, geng2022p5}, generating a variety of multimodal data 
with rich product images~\citep{jin2023eclip}.
These images show the details of the products and help customers make informed decisions.
However, the inclusion of a variety of images raises crucial questions about their actual utility: 
\emph{do they consistently contribute to the product understanding and decision-making, 
or do they sometimes introduce redundancy or even confuse customers?}
%
%
These questions need to be addressed through inspecting a variety of e-commerce visual data. 
However, existing multimodal e-commerce datasets are not ideal for evaluating or stratifying the utilities from product images
due to their significant limitations in scale and structural design~\citep{jin2024shoppingmmlu, peng2024ecellm}. 
Meanwhile, there is a lack of principled methods that can effectively utilize visual content for e-commerce applications, 
particularly in multimodal large language models (MLLMs).

%

To bridge the gap, we first introduce \dataset, a large-scale \underline{e}-\underline{com}merce 
\underline{m}ulti\underline{m}odal \underline{m}ultitask \underline{u}nderstanding dataset, 
designed to evaluate and benchmark visual utilities for e-commerce tasks. 
\dataset comprises 406,190 samples from real-world e-commerce applications 
with 8,989,510 product images spanning around 34 categories.
It is carefully crafted with rigorous data collection and structural design process to evaluate 
4 essential capabilities on modern e-commerce platforms 
with 8 tasks, such as question answering, query search, recommendation.
%
%

Building on top of that, 
we identify a \underline{v}ision-\underline{s}alient sub\underline{s}et (\hard) with 13,381 samples 
to 
explore model utilization of images in e-commerce applications.
Figure~\ref{fig:dataset_cases} shows a sample from \dataset identified as visual-salient.
Previous studies~\cite{chang2024survey, lu2022scienceqa} have shown that 
the inclusion of images does not consistently bring benefits over text alone.
%
To rigorously assess MLLMs' ability to use visual content, 
it is crucial to identify samples where text alone is insufficient 
but visuals provide added value.
%
%
To this end, we identify the specialized \hard subset within \dataset, 
on which at least 75\% of evaluated LLMs 
fail to provide correct responses for \hard set samples using text alone, as presented in Figure~\ref{fig:hard_identification}.
%
This consensus decision ensures that 
the identification is robust and model-grounded, 
and reflects the intrinsic limitations of product text.
With limited textual value in \hard, it serves as a robust visual utility benchmark by amplifying the models' strengths and weaknesses in visual utilization.



Preliminary analysis (Section~\ref{sec:init_analysis}) on \dataset of widely-used MLLMs
also underscores the strong promise of {\hard} set for 
benchmarking the visual utility in text-limited scenarios.
It reveals that
visual data does not always enhance model performance
-- indeed, images can sometimes negatively affect outcomes.
%
This is probably because MLLMs treat all images indistinguishably 
and do not differentiate their utilities.
Given that product images often contain potentially distracting visual details~\cite{hou2024bridging} that may impair model performance, there is an urgent need for a specialized method to effectively optimize multimodal learning with a variety of images in e-commerce.


%
Building upon the above observations, we further develop \method, 
a method that \underline{s}trategically \underline{u}tilizes \underline{m}ultiple \underline{e}-commerce \underline{i}mages
via predicting visual utilities before using them for downstream tasks.
\method evaluates visual utilities via a visual utility assessment component
to assess the utility of training sample images. 
%
Leveraging such training samples and their utility assessments, 
\method fine-tunes
a visual utility predictor
to predict \emph{helpful} images for the downstream tasks.
%
Using texts and images that are identified as helpful, we fine-tune a vision-salient MLLM to conduct downstream tasks.
%
%
Comprehensive experiments show that \method achieves the best performance compared with various language models, 
including Claude 3.5~\citep{claude3.5sonnet}, 
highlighting its strong capability for multimodal e-commerce.
%
Our data and code are available in \href{https://github.com/ninglab/EcomMMMU}{https://github.com/ninglab/EcomMMMU}.
\section{Related Work}
\label{sec:related}

\paragraph*{Evaluation of MLLMs}
The rise of the digital economy makes e-commerce a pivotal part of everyday life. The 
datasets aiming to develop and evaluate e-commerce methods are emerging~\citep{liu2016deepfashion, liu2023mep}. 
\citeauthor{reddy2022shopping} introduces a shopping query dataset to measure and improve the search quality on e-commerce platforms. Amazon-M2~\citep{jin2024amazonm2} introduces an e-commerce dataset with user sessions to help enhance personalization.
\citeauthor{li2024ecomgpt} and \citeauthor{peng2024ecellm} propose instruction-tuning datasets for e-commerce and utilize them to fine-tune e-commerce LLMs. ShoppingMMLU~\citep{jin2024shoppingmmlu} proposes a multi-task online shopping
benchmark dataset to evaluate the abilities of LLMs as shop assistants. MMECInstruct~\citep{ling2024caslie} introduces a multimodal instruction dataset for e-commerce applications. However, existing datasets for e-commerce models lack the scale of visual content to evaluate the inherent comprehension ability of models with a variety of images. In contrast, \dataset is a large-scale multitask multimodel understanding dataset with a variety of images and the specialized \hard subset. 

\paragraph*{E-commerce Foundation Models}
Recent progress in multimodal foundation models~\citep{radford2021clip,liu2023llava} has greatly advanced the modeling of vision and language. 
Models such as Claude 3.5 Sonnet~\citep{claude3.5sonnet} provide robust reasoning capabilities, while Phi-vision~\citep{abdin2024phi3} offers efficient, lightweight architectures suitable for various applications. Qwen-VL~\citep{Qwen-VL} specializes in detailed image comprehension, and Llava-interleave~\citep{li2024llava-next-interleave} effectively manages complex scenarios by interleaving multiple visual inputs, enriching contextual understanding. 
These advancements encourage the emergence of domain-specific models~\citep{chia2022fashionclip, jin2023eclip} designed to address the distinct needs of e-commerce environments. 
eCeLLM~\citep{peng2024ecellm} utilizes a domain-specific instruction dataset to improve performance. CASLIE~\citep{ling2024caslie} further refines this with captioned image inputs, emphasizing the need for effective utilization of multimodal data.
\method introduces domain-specific techniques to solve e-commerce applications to effectively utilize a variety of product images in e-commerce.

\section{\dataset Dataset}
\label{sec:dataset}

\begin{table}[ht]
  \centering
    \setlength{\tabcolsep}{1pt}
  \begin{footnotesize}
  \begin{threeparttable}
      \begin{tabular}{
	@{\hspace{0pt}}l@{\hspace{1pt}}
	@{\hspace{0pt}}c@{\hspace{0pt}}
	@{\hspace{0pt}}c@{\hspace{1pt}}      
	@{\hspace{2pt}}c@{\hspace{2pt}}      	
	@{\hspace{1pt}}r@{\hspace{0pt}}      
	}
      \toprule
{\textbf{Mod.}} & {\textbf{Dataset}} & \textbf{Div.} & \textbf{{VS}}  &  \textbf{Size} \\
	 \midrule
       \multirow{5}{*}{Text}
       & Amazon-M2~\citep{jin2024amazonm2} & \xmark & \xmark & 3.6M \\
       & Shopping Query~\citep{reddy2022shopping} & \xmark & \xmark & 130K \\
       & EcomInstruct~\citep{li2024ecomgpt} & \cmark & \xmark & 2.6M \\
       & ECInstruct~\citep{peng2024ecellm} & \cmark & \xmark & 117K \\
       & Shopping MMLU~\citep{jin2024shoppingmmlu} & \cmark & \xmark & 11K \\
       \midrule
       Text \&
       & MMECInstruct~\citep{ling2024caslie} & \cmark & \xmark & 75K \\
       Image & \dataset (ours) & \cmark & \cmark & 406K \\
      \bottomrule
      \end{tabular}
  \end{threeparttable}
  \end{footnotesize}
  \caption{Comparison between \dataset and relevant e-commerce datasets. ``Mod.'' denotes the type of data modalit(ies) in the dataset. ``Div.'' denotes whether the dataset contains diverse tasks. ``VS'' denotes whether the dataset contains identified visual-salient samples. ``Size'' denotes the number of data samples in each dataset.}
  \label{tbl:data_comp}
\end{table}

E-commerce platforms are rich in multimodal product information with a variety of images.
To systematically study the role of these images, 
we curate \dataset, in which each product is accompanied by at least 8 images along with textual contents (e.g., product titles, user reviews).
%
The dataset serves two primary purposes: 
\textbf{(1)} 
to determine which images in what context provided added value to multimodal learning;
\textbf{(2)} 
to benchmark model capabilities of utilizing a variety of visual information for e-commerce tasks.
\dataset contains a dedicated subset in which texts provide inadequate information, and visuals could bring improvement for tasks.
We compare \dataset with other e-commerce datasets in Table~\ref{tbl:data_comp}.
%

\subsection{Multi-image Vision-language Data}
\label{sec:multiimg_data}
We construct \dataset from diverse, real-world e-commerce data sources, carefully curating multimodal content to ensure realistic, rich, and high-quality data.
%
%
Each sample in \dataset comprises 
\textbf{(1)} multiple product images (i.e., visual content), including a designated main image as the default display on e-commerce platforms, 
and multiple others provided by producers or customers; 
\textbf{(2)} various texts (i.e., textual content) such as product descriptions (e.g., titles, categories, brands), and user-generated content 
(e.g., reviews, questions, queries); and 
\textbf{(3)} structured instructions tailored for real-world tasks over visual and textual content; and 
\textbf{(4)} ground-truth ``response" over each sample for specific e-commerce tasks.
\dataset is fundamentally different from text-only instruction datasets such as EcomInstruct~\citep{li2024ecomgpt} and 
single-image multimodal dataset such as MMECInstruct~\citep{ling2024caslie}.
%
A key characteristic of \dataset is its \hard subset, which is designed to rigorously assess the MLLMs’ abilities in utilizing product images (Section~\ref{sec:hard_set_data}).

\subsection{Data Collection}
\label{sec:data_collection}

\begin{table}[!ht]
  \centering
  \begin{footnotesize}
  \begin{threeparttable}
      \begin{tabular}{
	@{\hspace{0pt}}l@{\hspace{0pt}}
	@{\hspace{2pt}}r@{\hspace{2pt}}
	@{\hspace{2pt}}r@{\hspace{2pt}}
	@{\hspace{2pt}}r@{\hspace{2pt}}
	@{\hspace{2pt}}r@{\hspace{2pt}}
	@{\hspace{2pt}}r@{\hspace{2pt}}
	@{\hspace{2pt}}r@{\hspace{0pt}}
      }
      \toprule
       \textbf{Tasks} & \textbf{\#Train.} & \textbf{\#Valid.} & \textbf{\#\GTS} & \textbf{\#\VSS} & \textbf{\hspace{-2pt}\#Samples} & \textbf{\#Images}\\
	 \midrule
       \AP & 63,525 & 7,969 & 7,899 & 1,541 & 79,393 & 749,518 \\
       \QA & 9,150 & 1,152 & 1,105 & 406 & 11,407 & 107,070 \\
       \CP & 46,229 & 2,000 & 2,000 & 865 & 50,229 & 2,891,022 \\
       \SR & 46,211 & 2,000 & 2,000 & 1,844 & 50,211 & 2,896,026 \\
       \MPC & 53,420 & 8,199 & 8,073 & 2,954 & 69,692 & 627,918 \\
       \PSI & 53,420 & 8,199 & 8,090 & 3,253 & 69,709 & 628,086 \\
       \PRP & 29,158 & 3,744 & 3,671 & 1,940 & 36,573 & 683,660 \\
       \SA & 31,181 & 3,897 & 3,898 & 578 & 38,976 & 406,210 \\
       \midrule
       total & 332,294 & 37,160 & 36,736 & 13,381 & 406,190 & 8,989,510 \\
      \bottomrule
      \end{tabular}
  \end{threeparttable}
  \end{footnotesize}
  \caption{Summary of the \dataset dataset. ``\#Train.'', ``\#Valid.'', ``\#\GTS'' and ``\#\hard'' denote the number of samples in the training, 
  validation, general test set and vision-salient subset, respectively. 
  ``\#Samples'' denotes the total number of data samples for each task. ``\#Images'' denotes the total number of images for each task.}
  \label{tbl:data_summary}
\end{table}

\dataset is constructed by aggregating products from the real-world e-commerce platforms across diverse product categories. 
%
To ensure anonymity, all user IDs are removed. 
Additionally, low-quality images (e.g., with a low resolution) are filtered out to maintain the informativeness of the visual data.
%
As shown in 
Figure~\ref{fig:pie_dataset} in the Appendix, 
\dataset covers eight real-world e-commerce tasks: 
\textbf{(1)} question answerability prediction (\AP) and 
\textbf{(2)} binary question answering (\QA) for shopping question perception;
\textbf{(3)} click-through prediction (\CP) and 
\textbf{(4)} sequential recommendation (\SR) for user behavior alignment;
\textbf{(5)} multi-class product classification (\MPC) and 
\textbf{(6)} product substitute identification (\PSI) for query-product perception; and
\textbf{(7)} product relation prediction (\PRP) and 
\textbf{(8)} sentiment analysis (\SA) for shopping concept understanding.
The collected data is split into training, validation, and test (\GTS) sets per task with an 8:1:1 ratio.
The sizes for each set and a specialized \hard subset (Section~\ref{sec:hard_set_data}) are
shown in Table~\ref{tbl:data_summary}.
%
%
The task definitions, collection details, and detailed dataset statistics are presented in Appendix~\ref{sec:appendix:data}. 
The instruction templates for the tasks are detailed in Appendix~\ref{sec:appdix:templates}.

\subsection{Selection of \hard Set}
\label{sec:hard_set_data}

While \dataset covers a wide variety of multimodal e-commerce tasks with both textual and visual information, 
the contribution of visual information varies across samples.
Some products benefit significantly from accompanying images, particularly 
when textual descriptions cannot provide sufficient information for correct prediction~\citep{ma2022multimodal}, 
while others can be effectively understood using text alone.
%
Identifying samples where text alone is insufficient and visual information adds meaningful value 
is crucial for evaluating the real contribution of visual content and for benchmarking 
how effectively MLLMs leverage such visual content.
To systematically identify such samples,
we adopt a model-consensus-based approach inspired by~\citet{chen2024mmstar}, 
and construct a vision-salient subset (\VSS) from the general test set (\GTS).
 

Specifically, we identify a sample as \VSS if at least 75\% (6 out of 8) strong LLMs
fail to answer correctly (i.e., LLM's output cannot match the ground-truth response on the sample) 
using only its textual content as input under 2-shot in-context prompting. 
%
%
Thus, the consensus decision across multiple LLMs ensures that the identification is robust and model-grounded, and reflects 
the inherent limitations of textual information alone. 
We provide a human evaluation 
on \hard in Appendix~\ref{sec:appdix:vc_set} to demonstrate the effectiveness of our subset selection strategy.
%
%
%

\subsection{Preliminary Analysis on Existing MLLMs}
\label{sec:init_analysis}


With \dataset, we perform a zero-shot evaluation on all tasks for prominent MLLMs to assess 
their capabilities of multimodal e-commerce applications.
%
We specifically assess how effectively these models utilize visual information by comparing their performance 
using: 
\textbf{(1)} textual content only, 
\textbf{(2)} main product image (i.e., the default display) and textual content,
\textbf{(3)} all available product images and textual content.
The MLLMs include
Claude 3.5~\citep{claude3.5sonnet}, Phi-vision~\citep{abdin2024phi3}, Qwen2.5-VL~(7B)~\citep{qwen2.5-VL}, and Llava-interleave~\citep{li2024llava-next-interleave}
on both \GTS
and \VSS subset 
using accuracy across all samples.
%
We have the following key insights from Table~\ref{tbl:init_analysis}.

\begin{table}[ht]
  \centering
  \begin{footnotesize}
  \begin{threeparttable}
      \begin{tabular}{
	@{\hspace{0pt}}l@{\hspace{2pt}}
	@{\hspace{2pt}}l@{\hspace{2pt}}
	@{\hspace{5pt}}c@{\hspace{5pt}}
	@{\hspace{5pt}}c@{\hspace{0pt}}
      }
      \toprule
       \textbf{MLLM} & \textbf{~~~~Modalities} & \textbf{\GTS} & \textbf{\VSS} \\
	 \midrule
       \multirow{3}{*}{Claude 3.5} 
       & text & 0.488 & 0.175 \\
       & text + main image & 0.551 & 0.261 \\
       & text + multiple images  & 0.349 & 0.178 \\
       \midrule
       \multirow{3}{*}{Phi-vision} 
       & text & 0.481 & 0.186 \\
       & text + main image & 0.480 & 0.184 \\
       & text + multiple images & 0.455 & 0.156 \\
       \midrule
       \multirow{3}{*}{Qwen2.5-VL} 
        & text & 0.510 & 0.246 \\
       & text + main image & 0.515 & 0.249 \\
       & text + multiple images & 0.163 & 0.157 \\
       \midrule
       \multirow{3}{*}{Llava-interleave} 
       & text & 0.469 & 0.226 \\
       & text + main image & 0.447 & 0.227 \\
       & text + multiple images  & 0.333 & 0.164 \\
      \bottomrule
      \end{tabular}
  \end{threeparttable}
  \end{footnotesize}
  \caption{Preliminary results on MLLMs. Performance is measured in accuracy.}
  \label{tbl:init_analysis}
\end{table}

\textbf{(1) Benchmarking the Utility of \hard set:}
Comparing the performance on \GTS with that on \hard subset using text-only input, 
all models show a significant performance drop (e.g., 0.488 to 0.175 for Claude 3.5). 
This discrepancy underscores the rationale of our consensus-based \hard design, 
in which samples carry inadequate text information in general.
%
Meanwhile, Claude 3.5 shows a 49.14\% improvement 
by adding the main image to the text input
on the \hard set, significantly higher than that on \GTS (12.91\% improvement).
Serving as the most powerful model among the evaluated MLLMs
(based on Table~\ref{tbl:init_analysis} and
Section~\ref{sec:overall_eval}),
the results of Claude 3.5 highlight the strong promise of \hard to benchmark and evaluate visual utilities 
as the potential improvement over \hard due to visuals can be more noticeable than over average data. 
They also indicate that \hard is effective in 
benchmarking 
model effectiveness in utilizing visual information 
when texts are insufficient. 
%



\textbf{(2) Underutilization of Visuals:} 
Visual information does not always lead to performance gains across MLLMs. 
Among the four MLLMs, when adding the main product image to the text input, 
Llava-interleave exhibits noticeable performance degradation in \GTS,  
dropping from 0.469 to 0.447; 
Phi-vision and Qwen2.5-VL show very limited improvement, and only Claude3.5 is enhanced by visuals. 
In addition, incorporating multiple images consistently impedes MLLMs' performance 
compared with using text data.
Similar observations hold true for \VSS.
These indicate potential obscurity from misleading images 
and underutilization of visual content.

%
%

\textbf{(3) Negative Impact of Redundant Images:} 
%
Compared to using only the main image, performance drops significantly across four MLLMs 
when multiple images are included, in both \GTS and \hard. 
This phenomenon is most notable in Qwen2.5-VL, whose accuracy on \GTS falls from 0.510 to 0.157. 
This may suggest that a variety of images together may introduce redundant, conflicting, or distracting visual information 
that hampers model comprehension. 
These observations validate the necessity and effectiveness of \hard set, and reveal a key insight: while visual content often contains rich product information, using it indiscriminately can hinder performance. This underscores the need for specialized strategies to optimize multimodal learning in e-commerce.

\section{\method: Strategic Utilization of Multiple E-commerce Images}
\label{sec:method}

\begin{figure}[ht]
    \centering
    \includegraphics[trim=44 81 44 81, clip, width=\linewidth]{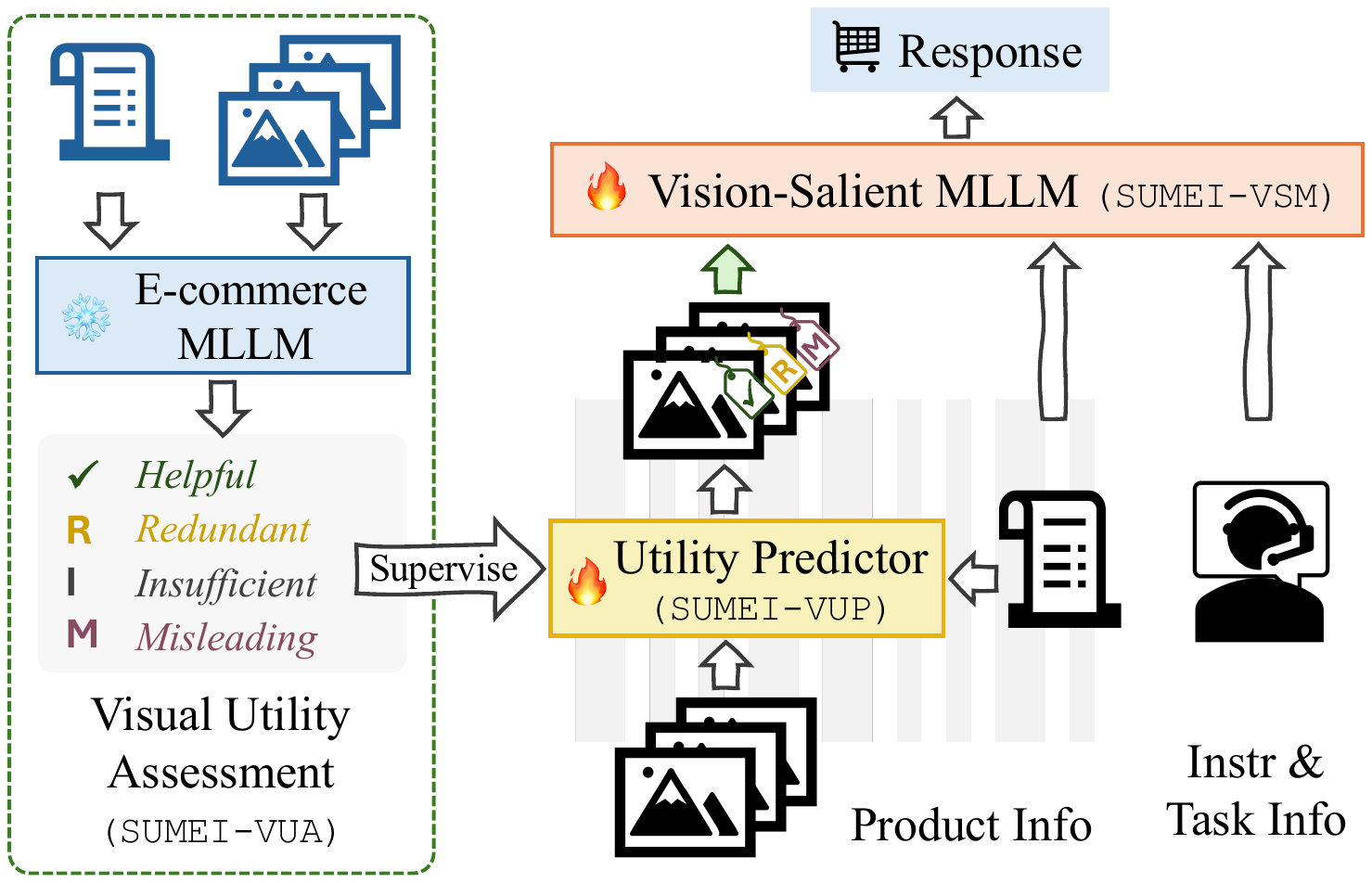}  
    \caption{Overall workflow of \method.}
    \label{fig:framework}
\end{figure}
 
As discussed in Section~\ref{sec:init_analysis}, although images may bring improvement, incorporating all the images without discrimination can undermine performance.
This emphasizes the need for a method that can strategically and selectively utilize product images for e-commerce.
%
In this section, we introduce \method, a data-driven
method that strategically distinguishes the utilities of images for the tasks and dynamically integrates them 
in the downstream model based on their utilities.
%
Figure~\ref{fig:framework} shows the overall scheme of \method.

\subsection{Visual Utility Assessment (\VUA)}
\label{sec:pseudo_labeling}


Whether an image is useful or not is inherently subjective and context-dependent. 
It can vary across tasks and depend heavily on accompanying textual information. 
This calls for a data-driven, model-grounded approach to assessing visual utilities. 
Moreover, proactive assessment before using images  
can enable informed decisions on whether and how to incorporate visual content. 
%
Unfortunately, reliable visual utility measurements are typically unavailable, and 
it is costly, if ever possible, to manually annotate at scale. 
Thus, \method adopts an automated assessment strategy, denoted as \VUA (\underline{v}isual \underline{u}tility \underline{a}ssessment), 
and infers visual utilities based on performance disparity
on downstream tasks~\citep{sohn2020fixmatch}, 
enabling scalable and automated assessment in a task- and context-aware manner.

%

Specifically, during the training stage, 
\method evaluates the performance on each training sample 
using an e-commerce MLLM~\cite{liu2024llava-next} 
to perform the target task (Section~\ref{sec:data_collection}) 
with two types of inputs:
\textbf{(1)} multimodal input: text combined with one product image that will be assessed, and \textbf{(2)} unimodal input with text only.
The performance differences characterize visual utilities into four categories:

\noindent
\textbf{(1)} \emph{Helpful}, when the
model response is correct with multimodal input but incorrect with text-only input, highlighting essential visual utilities.

\noindent
\textbf{(2)} \emph{Redundant},  when
both multimodal and text-only inputs yield the correct response, indicating that visual information adds no further value. 

\noindent
\textbf{(3)} \emph{Insufficient}, when neither multimodal nor text-only inputs allow the model to achieve a correct response.

\noindent
\textbf{(4)} \emph{Misleading}, when 
images do not lead to a correct response while text-only inputs do. 

Examples of visual utility assessment are presented in Figure~\ref{fig:dataset_cases}. 
%

\subsection{Visual Utility Prediction (\VUP)}
\label{sec:image_verifier}

During training, we fine-tune a visual utility predictor, 
denoted as \VUP, using the training samples and 
their corresponding visual utilities assessed by \VUA. 
\VUP will be used during inference to predict the utility of new sample images, 
when the correct response over the new sample
is unknown and thus \VUA is not applicable. 
Based on \VUP's results, \method selects the \emph{helpful} images for the downstream tasks. 
%


Specifically, \VUP learns from the textual content, multiple images, and their assessed visual utilities
(as image ``labels") of training samples
to predict visual utilities for new samples. 
We fine-tune a general MLLM~\citep{li2024llava-next-interleave} as \VUP, 
since the visual assessment by \VUA is depicted in a textual form, 
each carrying specific semantic meanings (\emph{helpful}, \emph{redundant}, \emph{misleading}, \emph{insufficient}), 
inherently aligned with the strengths of language modeling. 
In addition, employing MLLMs allows the model to leverage its world knowledge to understand textual conditions and visual contents, 
enhancing the learning of \VUP. 
Serving as a core component of \method, \VUP enables dynamic, 
scalable, context-aware visual utility prediction, and thus, the corresponding image selection, before downstream model training.
%

\subsection{Vision-Salient MLLM (\VSM)}
\label{sec:model_training}

With \VUP, we fine-tune a generalist vision-salient MLLM, denoted as \VSM, within the \method 
for the downstream tasks. 
First, \method takes textual content alongside multiple images of each sample as input, 
and applies \VUP to select the 
\emph{helpful} image(s) for the sample.
In case multiple images are predicted as \emph{helpful}, \method randomly selects one
to avoid the potential latent effects that reduce their combined utility.
If no \emph{helpful} image can be identified, \method will only use the textual information of the sample; otherwise, 
the textual information and the \emph{helpful} image. 
An MLLM will be fine-tuned within \method using such information 
into \VSM. 
Note that half of the training and validation sets of \dataset are used for \VUP training, and the other halves are 
used for \VSM fine-tuning.

%

\section{Experiments}
\label{sec:experiment}

We conduct a systematic evaluation of \dataset. 
We present the experiment setups in Section~\ref{sec:exp_setup}, 
the overall analysis of \GTS in Section~\ref{sec:overall_eval}, 
the analysis on \VSS in Section~\ref{sec:hard_eval}, and the ablation studies in Section~\ref{sec:ablation}
and Section~\ref{sec:appendix:vua}. 

\subsection{Experimental Setups}
\label{sec:exp_setup}

We employ zero-shot evaluation that aligns with real-world scenarios where customers typically input queries directly 
without few-shot examples. 
For LLMs as baseline models, we evaluate them with text-only product information; 
for MLLM baselines, we use the main product image and text; 
for our proposed \method, 
we use multiple images accompanied with textual data.
%
Open-source models employ the Huggingface~\citep{wolf-etal-2020-transformers} checkpoints.
Closed-source models employ their official APIs. 
We utilize Llava-interleave-qwen-7B~\citep{li2024llava-next-interleave} and CASLIE-M~\citep{ling2024caslie} as fine-tuned downstream MLLM 
backbones in \method, denoted as \methodL and \methodC, respectively. The model size and budget are presented in Appendix~\ref{sec:appendix:budget}.

\paragraph{Baseline Models}
\textbf{(1) General LLMs:} We evaluate general LLMs including Mistral-7B-v0.3~\citep{jiang2023mistral}, Ministral~\citep{mistral2024ministral}, Qwen2.5-7B, Qwen2.5-14B~\citep{qwen2.5}, Gemma2-9B, Gemma2-27B~\citep{gemma2}, Phi3.5~\citep{abdin2024phi3}, and Phi4~\citep{abdin2024phi4}. 
\textbf{(2) E-Commerce LLMs:} Powerfull e-commerce LLMs such as eCeLLM-M and eCeLLM-L~\citep{peng2024ecellm} are evaluated. 
\textbf{(3) General MLLMs:} We assess general MLLMs including Claude 3.5~\citep{claude3.5sonnet}, Phi-vision~\citep{abdin2024phi3}, Qwen2.5-VL (7B)~\citep{qwen2.5-VL}, and Llava-interleave (Llava-ITL for short in Section~\ref{sec:experiment})~\citep{li2024llava-next-interleave}.
\textbf{(4) E-commerce MLLMs:} We evaluate CASLIE-M and CASLIE-L~\citep{ling2024caslie}, the leading MLLM series, which also selectively integrate visual information but only in caption format.

\paragraph{Evaluation Metrics}
We utilize a comprehensive set of metrics on each task for extensive evaluations.
The complete results are listed in Appendix~\ref{sec:appdix:full_result}. For clarity, here we present the performance of the primary metrics for each task, including accuracy (for \CP, \MPC, and \QA), F1-score (for \AP and \PSI), macro F1-score (for \PRP and \SA), and Recall@1 (for \SR) to accommodate the emphasis of each task. 
In addition, to fairly compare models across diverse tasks with different metrics, we rank each model per task and compute the average rank ($\text{R}_{\text{avg}}$) as an overall performance indicator to avoid metric scale inconsistencies. For a model $M_i$,
%
\begin{equation*}
\text{R}_{\text{avg}}(M_i) = \frac{1}{T} \sum\nolimits_{t=1}^{T} R_{i,t},
\end{equation*}
where $R_{i,t}$ is the rank of $M_i$'s performance in task t, 
and $R_{i,t} = 1$ indicates model $M_i$ achieve the best performance in task $i$. $T$ is the total number of tasks. Lower $\text{R}_{\text{avg}}$ indicates better performance. 

\subsection{Overall Evaluation on \GTS}
\label{sec:overall_eval}
\begin{table}[ht]
  \centering
  \begin{scriptsize}
  \begin{threeparttable}
      \begin{tabular}{
	@{\hspace{0pt}}l@{\hspace{2pt}}
	@{\hspace{2pt}}c@{\hspace{2pt}}
	@{\hspace{2pt}}c@{\hspace{2pt}}
	@{\hspace{2pt}}c@{\hspace{2pt}}
	@{\hspace{2pt}}c@{\hspace{2pt}}
	@{\hspace{2pt}}c@{\hspace{2pt}}
	@{\hspace{2pt}}c@{\hspace{2pt}}
	@{\hspace{2pt}}c@{\hspace{2pt}}
	@{\hspace{2pt}}c@{\hspace{2pt}}
	|@{\hspace{2pt}}r@{\hspace{0pt}}
      }
      \toprule        
       \multirow{1}{*}{\textbf{Model}}
       & {\AP} & {\QA} & {\CP} & {\SR} & {\MPC} & {\PSI} & {\PRP} & {\SA} & \textbf{R}$_\text{avg}$ \\
       \midrule
       
\multicolumn{10}{c}{\textit{General and E-commerce LLMs}} \\
\cmidrule(){1-10}
Mistral-v0.3 & 0.747 & 0.527 & 0.570 & 0.002 & 0.660 & 0.373 & 0.297 & 0.570 & 9.750 \\
Ministral       & 0.357 & 0.261 & 0.532 & 0.000 & 0.645 & 0.269 & 0.187 & 0.416 & 16.375 \\
Qwen2.5-7B      & 0.703 & 0.551 & 0.511 & 0.126 & 0.624 & 0.212 & 0.165 & 0.571 & 13.375 \\
Qwen2.5-14B     & 0.542 & 0.581 & 0.592 & 0.129 & 0.509 & 0.314 & 0.267 & 0.488 & 10.875 \\
Gemma2-9B       & 0.686 & 0.555 & 0.589 & 0.072 & 0.611 & 0.316 & 0.201 & 0.562 & 11.625 \\
Gemma2-27B      & 0.775 & 0.570 & 0.600 & 0.100 & 0.657 & 0.372 & 0.192 & 0.618 & 8.500 \\
Phi3.5          & 0.780 & 0.430 & 0.585 & 0.073 & 0.607 & 0.334 & 0.187 & 0.226 & 12.625 \\
Phi4            & 0.580 & 0.505 & 0.587 & 0.003 & 0.669 & 0.370 & 0.247 & 0.581 & 10.125 \\
\cmidrule(){2-10}
eCeLLM-M        & 0.861 & 0.351 & 0.549 & 0.209 & 0.713 & 0.373 & 0.518 & 0.662 & 6.125 \\
eCeLLM-L        & 0.840 & 0.234 & 0.545 & 0.193 & 0.661 & 0.413 & 0.545 & 0.619 & 8.000 \\
\cmidrule(){1-10}

\multicolumn{10}{c}{\textit{General and E-commerce MLLMs}} \\
\cmidrule(){1-10}
Claude 3.5          & 0.761 & 0.540 & 0.584 & 0.149 & 0.685 & 0.361 & 0.265 & 0.663 & 7.000 \\
Phi-vision          & 0.538 & 0.312 & 0.544 & 0.100 & 0.658 & 0.320 & 0.243 & 0.492 & 13.375 \\
Qwen2.5-VL          & 0.730 & 0.483 & 0.002 & 0.000 & 0.658 & 0.282 & 0.295 & 0.635 & 12.000 \\
Llava-ITL   & 0.727 & 0.329 & 0.573 & 0.046 & 0.645 & 0.339 & 0.193 & 0.483 & 13.000 \\
\cmidrule(){2-10}
CASLIE-M & 0.863 & 0.438 & 0.582 & 0.219 & \textbf{0.723} & 0.464 & 0.553 & 0.660 & 4.375 \\
CASLIE-L & 0.824 & 0.329 & 0.552 & 0.143 & 0.674 & 0.342 & 0.470 & 0.625 & 8.375 \\
\cmidrule(){1-10}

\multicolumn{10}{c}{\textit{Ours}} \\
\cmidrule(){1-10}
\methodL & 0.823 & 0.577 & 0.607 & 0.173 & 0.693 & 0.457 & \textbf{0.576} & 0.663 & 3.250 \\
\methodC & \textbf{0.872} & \textbf{0.596} & \textbf{0.649} & \textbf{0.231} & 0.721 & \textbf{0.525} & 0.555 & \textbf{0.681} & \textbf{1.250} \\

      \bottomrule
      \end{tabular}
  \end{threeparttable}
  \end{scriptsize}
  \caption{Performance comparison on \GTS set. The best performance on each task is in \textbf{bold.}}
  \label{tbl:general_res}
\end{table}


Table~\ref{tbl:general_res} presents the model performance on the general test set \GTS. 

First, \textbf{\textit{\methodC performs the best across all the tasks (R$_\text{avg}$ = 1.250) compared with both general and e-commerce LLMs and MLLMs}}.
This is attributed to our strategic design of \method to 
effectively utilize a variety of visual information, 
while LLMs cannot utilize images, and MLLMs cannot distinguish visual utility before using them.
\methodL is the second-best method (R$_\text{avg}$ = 3.250), 
further indicating the superiority of \method in predicting and leveraging image utilities 
for downstream tasks. 

Second, 
\textbf{\textit{MLLMs show marginal gains over LLMs, but do not consistently outperform them.}}
In several cases, such as Phi4 vs Phi-vision (similar series model),
LLMs demonstrate comparable or even superior performance. 
The fact that such inconsistencies can be systematically observed across diverse tasks in \GTS 
indicates and validates a general trend that indiscriminate use of visual data does not lead to 
performance improvement, which is consistent with the findings in the literature~\cite{chang2024survey}.
%
Meanwhile, 
by including large-scale, diverse, and realistic multimodal e-commerce scenarios with 
rich visual content, our benchmarking dataset effectively reflects nuanced multimodal behaviors, 
making it a rigorous and informative benchmark for e-commerce applications. 



Third, 
\textbf{\textit{e-commerce models exhibit robust performance with clear advantages over general models in addressing domain-specific challenges,
and \method brings further improvements. }}
For example, though CASLIE-M adopts a selective visual incorporation method by zero-shot evaluation,
\methodC achieves better results by utilizing explicit visual utility assessment to train the predictor.
This is particularly evident in complex tasks such as \PSI ({0.525 vs. 0.464}), 
where visual understanding is essential.
Such performance gain highlights the effectiveness 
of \method
to utilize a variety of images for e-commerce.


\subsection{Overall Evaluation on \hard}
\label{sec:hard_eval}
\begin{table}[!h]
  \centering
  \begin{scriptsize}
  \begin{threeparttable}
      \begin{tabular}{
	@{\hspace{0pt}}l@{\hspace{2pt}}
	@{\hspace{2pt}}c@{\hspace{2pt}}
	@{\hspace{2pt}}c@{\hspace{2pt}}
	@{\hspace{2pt}}c@{\hspace{2pt}}
	@{\hspace{2pt}}c@{\hspace{2pt}}
	@{\hspace{2pt}}c@{\hspace{2pt}}
	@{\hspace{2pt}}c@{\hspace{2pt}}
	@{\hspace{2pt}}c@{\hspace{2pt}}
	@{\hspace{2pt}}c@{\hspace{2pt}}
	|@{\hspace{2pt}}r@{\hspace{0pt}}
      }
      \toprule        
       \multirow{1}{*}{\textbf{Model}}
       & {\AP} & {\QA} & {\CP} & {\SR} & {\MPC} & {\PSI} & {\PRP} & {\SA} & \textbf{R}$_\text{avg}$ \\
       \midrule
\multicolumn{10}{c}{\textit{General and E-commerce MLLMs}} \\
\cmidrule(){1-10}
Claude 3.5          & 0.664 & 0.283 & 0.519 & 0.105 & 0.443 & 0.087 & 0.066 & 0.156 & 5.000 \\
Phi-vision          & 0.273 & 0.192 & 0.523 & 0.075 & 0.418 & 0.030 & 0.036 & 0.186 & 6.000 \\
Qwen2.5-VL          & 0.631 & 0.249 & 0.000 & 0.029 & 0.369 & 0.015 & 0.256 & 0.083 & 7.250 \\
Llava-{ITL}    & 0.742 & 0.175 & 0.360 & 0.053 & 0.438 & 0.069 & 0.017 & 0.208 & 6.000 \\
\cmidrule(){2-10}
CASLIE-M & 0.824 & 0.372 & 0.491 & 0.191 & \textbf{0.506} & 0.317 & 0.362 & 0.208 & 2.625 \\
CASLIE-L & 0.781 & 0.313 & 0.459 & 0.119 & 0.399 & 0.349 & 0.291 & 0.183 & 4.375 \\
\cmidrule(){1-10}

\multicolumn{10}{c}{\textit{Ours}} \\
\cmidrule(){1-10}
\methodL  & \textbf{0.948} & \textbf{0.384} & 0.449 & 0.136 & 0.438 & 0.327 & \textbf{0.393} & \textbf{0.233} & 2.500 \\
\methodC & 0.848 & 0.305 & \textbf{0.529} & \textbf{0.197} & 0.481 & \textbf{0.418} & 0.362 & 0.210 & \textbf{1.875} \\

      \bottomrule
      \end{tabular}
  \end{threeparttable}
  \end{scriptsize}
  \caption{Performance comparison on the \hard set. The best performance on each task is in \textbf{bold.}}
  \label{tbl:hard_res}
   \vspace{-4pt}
\end{table}

Table~\ref{tbl:hard_res} presents the performance on the \hard set.
We do not include the evaluation of LLMs as the \hard set is selected by those LLMs. In general, \method exhibits advantages in the \hard set,
particularly \methodC, which achieves the best R$_\text{avg}$ (1.875) and outperforms general MLLMs across all tasks. 
Compared with the specialized e-commerce MLLMs (CASLIE-M and CASLIE-L), which also leverage selective multimodal learning strategies, our method consistently achieves superior performance across most tasks. 
The performance gaps suggest that \method mitigates visual noise, selects and utilizes 
helpful images more effectively, while simply adding visual content does not guarantee improvement. 
This superiority underscores the avail of {\method}, leading to promising results on {\hard}. 

The results also validate \dataset, a very robust benchmarking dataset with \hard subset, in assessing 
the practical utilities of visual contents to current MLLMs. 
Overall, baseline MLLMs exhibit lower performance on the \hard set compared to 
\method. 
Notably, general MLLMs such as Qwen2.5-VL struggle significantly on this set, 
achieving only 0.015 in \PSI and 0.083 in \SA. 
These suboptimal performances suggest 
they may be experiencing difficulties in taking advantage of visual information 
to handle cases where visuals are crucial for correct predictions. 
%
%
Such weakness is explicitly exposed under the evaluation on \hard.
The \hard set accomplishes its mission by 
explicitly disentangling scenarios in which visual content is helpful from others, 
thus, amplifying the visibility of models' strengths and weaknesses in visual utilization. 
\method's robust performances on the \hard set signify that 
the targeted design of \hard set enables 
a reliable benchmarking of multimodal competence in complex scenarios and 
highlights the practical limitations of existing MLLMs. 
%


\subsection{Ablation Study on \method}
\label{sec:ablation}
\begin{table}[!h]
  \centering
    \setlength{\tabcolsep}{1pt}
  \begin{scriptsize}
  \begin{threeparttable}
      \begin{tabular}{
	@{\hspace{0pt}}l@{\hspace{1pt}}
	@{\hspace{1pt}}l@{\hspace{6pt}}
	@{\hspace{6pt}}c@{\hspace{2pt}}
	@{\hspace{2pt}}c@{\hspace{2pt}}
	@{\hspace{2pt}}c@{\hspace{2pt}}
	@{\hspace{2pt}}c@{\hspace{2pt}}
	@{\hspace{2pt}}c@{\hspace{2pt}}
	@{\hspace{2pt}}c@{\hspace{2pt}}
	@{\hspace{2pt}}c@{\hspace{2pt}}
	@{\hspace{2pt}}c@{\hspace{2pt}}
	@{\hspace{2pt}}c@{\hspace{0pt}}
      }
      \toprule        
       \textbf{Set} & \multirow{1}{*}{\textbf{Base}} & \textbf{Img.} & \AP & \QA & \CP & \SR & \MPC & \PSI & \PRP & \SA \\
       \midrule
       
\parbox[t]{2mm}{\multirow{7}{*}{\rotatebox[origin=c]{90}{\GTS}}}

& \multirow{3}{*}{\parbox[t]{2mm}{{\rotatebox[origin=c]{0}{Llava}}}} &  \emph{helpful} 
& 0.823 & 0.577 & 0.607 & 0.173 & 0.693 & 0.457 & \textbf{0.576} & 0.663 \\
&  &  main 
& 0.748 & 0.513 & 0.510 & 0.086 & 0.714 & 0.294 & 0.575 & 0.631 \\
&  &  multiple
& 0.163 & 0.304 & 0.493 & 0.063 & 0.706 & 0.205 & 0.187 & 0.604 \\

\cmidrule(){2-11}
& \multirow{3}{*}{\parbox[t]{2mm}{{\rotatebox[origin=c]{0}{CASLIE}}}} &  \emph{helpful} 
& \textbf{0.872} & \textbf{0.596} & \textbf{0.649} & \textbf{0.231} & 0.721 & \textbf{0.525} & 0.555 & \textbf{0.681} \\
& &  main 
& 0.867 & 0.538 & 0.616 & 0.223 & \textbf{0.725} & 0.507 & 0.541 & 0.675 \\
& &  multiple
& 0.854 & 0.486 & 0.501 & 0.192 & 0.723 & 0.459 & 0.536 & 0.668 \\
\cmidrule(){2-11}
& \multicolumn{2}{c}{\emph{helpful} img \%} & 12.85 & \phantom{0}9.85 & 14.90 & 6.25 & 31.84 & 13.80 & 42.33 & 11.08 \\

\midrule
\parbox[t]{2mm}{\multirow{7}{*}{\rotatebox[origin=c]{90}{\hard}}}

& \multirow{3}{*}{\parbox[t]{2mm}{{\rotatebox[origin=c]{0}{Llava}}}} &  \emph{helpful} 
& \textbf{0.948} & 0.384 & 0.449 & 0.136 & 0.438 & 0.327 & \textbf{0.393} & \textbf{0.233} \\
& &  main 
& 0.522 & \textbf{0.391} & 0.376 & 0.068 & 0.465 & 0.309 & 0.369 & 0.206 \\
& &  multiple
& 0.022 & 0.165 & 0.346 & 0.050 & 0.415 & 0.090 & 0.017 & 0.143 \\

\cmidrule(){2-11}
& \multirow{3}{*}{\parbox[t]{2mm}{{\rotatebox[origin=c]{0}{CASLIE}}}} &  \emph{helpful} 
& 0.848 & 0.305 & \textbf{0.529} & \textbf{0.197} & 0.481 & \textbf{0.418} & 0.362 & 0.210 \\
& &  main 
& 0.810 & 0.288 & 0.509 & 0.179 & \textbf{0.486} & 0.403 & 0.342 & 0.189 \\
& &  multiple
& 0.815 & 0.177 & 0.304 & 0.157 & 0.485 & 0.309 & 0.301 & 0.180 \\
\cmidrule(){2-11}
& \multicolumn{2}{c}{ \emph{helpful} img \%}  & 21.52 & 17.35 & 19.05 & 12.52 & 43.72 & 24.07 & 56.56 & 19.63 \\

      \bottomrule
      \end{tabular}
      
  \end{threeparttable}
  \end{scriptsize}
  \caption{Ablation study on \method. ``Set'' denotes the split of \GTS or \VSS set. ``Base'' denotes the base MLLMs used in \method. ``Img.'' denotes the types of images that the MLLMs are fine-tuned with. ``\emph{helpful} img \%'' denotes the percentage of images that are identified as \emph{helpful}. The best performances of each task on the \GTS and \VSS are in \textbf{bold}.}
  \label{tbl:ablation}
\end{table}



Table~\ref{tbl:ablation} 
presents the results of \method contrasting against the same backbones without the visual utility prediction: 
direct fine-tuning MLLMs by textual content with the main image 
and with multiple available images, respectively.

First, \method using the \emph{helpful} image shows consistent improvement over the variants 
using the main image or multiple images across different base models and test sets.
The results suggest the complexity and difficulty for models to effectively utilize visuals in e-commerce tasks,
and demonstrate the effectiveness of \method in addressing such challenges.
%
Additionally, higher image usage percentages on the \hard set underline the increasing demands for visual content in this set.

Second, consistent with insights from preliminary analysis, 
direct fine-tuning using all images yields notably diminished performance 
across both \GTS and \hard. 
For instance, \methodL using multiple images has significantly lower scores 
in tasks \AP (0.163 \GTS, 0.022 \hard) compared to using the \emph{helpful} image (0.823 and 0.948). 
Similarly, \methodC experiences a performance decline by using multiple images compared with using the \emph{helpful} image, 
emphasizing the inefficiency of indiscriminate image integration.
However, {\method} achieves notable performance by 
assessing image utility, then selectively incorporating visuals. 
%
Thus, \method provides a principled and effective method for optimizing multi-image learning in e-commerce.


\subsection{Ablation Study on Image Assessment}
\label{sec:appendix:vua}

\begin{table}[!h]
  \centering
  \begin{footnotesize}
  \begin{threeparttable}
      \begin{tabular}{
	@{\hspace{0pt}}l@{\hspace{1.5pt}}
	@{\hspace{1.5pt}}c@{\hspace{1.5pt}}
	@{\hspace{1.5pt}}c@{\hspace{1.5pt}}
	@{\hspace{1.5pt}}c@{\hspace{1.5pt}}
	@{\hspace{1.5pt}}c@{\hspace{1.5pt}}
	@{\hspace{1.5pt}}c@{\hspace{1.5pt}}
	@{\hspace{1.5pt}}c@{\hspace{1.5pt}}
	@{\hspace{1.5pt}}c@{\hspace{1.5pt}}
	@{\hspace{1.5pt}}c@{\hspace{0pt}}
      }
      \toprule        
       \textbf{Img.} & \AP & \QA & \CP & \SR & \MPC & \PSI & \PRP & \SA \\
       \midrule
\it{h} & 0.823 & 0.577 & 0.607 & 0.173 & 0.693 & 0.457 & 0.576 & 0.663 \\
\it{h+r} & 0.830 & 0.571 & 0.596 & 0.168 & 0.671 & 0.450 & 0.569 & 0.675 \\
\it{h+r+i} & 0.825 & 0.532 & 0.577 & 0.163 & 0.615 & 0.394 & 0.570 & 0.652 \\
\it{h+r+i+m} & 0.734 & 0.296 & 0.526 & 0.078 & 0.609 & 0.283 & 0.536 & 0.631 \\

      \bottomrule
      \end{tabular}
  \end{threeparttable}
  \end{footnotesize}
  \caption{Ablation study of \VUA on \methodL on \GTS. 
  ``Img." denotes the types of images used for each task. 
  ``\emph{h}'', ``\emph{r}'', ``\emph{i}'', and ``\emph{m}'' denote the \emph{helpful}, \emph{redundant}, \emph{insufficient}
   and \emph{misleading} images, respectively.}
  \label{tbl:pseudo-labeling}
\end{table}


The visual utility assessment is performance-driven, 
and reflects the functional utility of images for the model on specific tasks. 
To validate the effectiveness of our assessment design, we conduct an ablation study 
on \methodL using different images assessed by \VUA on \GTS. 
%
As shown in Table~\ref{tbl:pseudo-labeling}, using only \emph{helpful} images yields 
the overall best results across most tasks. 
Adding \emph{redundant} or \emph{insufficient} images slightly reduces performance, 
while including \emph{misleading} images leads to substantial degradation. 
%
%
These results confirm that 
\VUA 
reliably separates helpful images from other images. 
%

\section{Conclusion}
\label{sec:conclusion}
In this work, we introduce \dataset, a large-scale multimodal multitask understanding dataset tailored to evaluate essential e-commerce applications, along with a \hard subset to assess models' abilities to use visual content. 
Through preliminary analysis, we demonstrate that the indiscriminate use of images can be detrimental to model performance. 
To overcome this, we propose \method, a method that predicts visual utilities before using them for downstream tasks, 
bringing improved performance and robustness across e-commerce applications. \dataset and \method offer a robust pathway for advancing multimodal learning in real-world e-commerce scenarios.

\section{Limitations}
Although \dataset is carefully constructed, some samples may contain noisy text-image pairs, which can affect model training and visual utility assessment results. 
To the best of our knowledge, perfect denoising of a large-scale dataset is impractical in current work~\cite{fang2023mol, peng2024ecellm}, but can be minimized by rigorous collection design. 
Our method depends on utility assessment generated from model output shifts~\citep{sohn2020fixmatch}. These assessments reflect model-specific behavior and may be unreliable when the base MLLM misinterprets inputs. As such, the effectiveness of \method is inherently tied to the reasoning stability of the underlying model. With the advances in future MLLMs, further improvement in \method could be expected.
Note that \method is tailored to e-commerce scenarios where multiple images depict a single product. This design limits generalization to domains with heterogeneous or ungrounded image sets, such as open-domain visual question answering~\citep{antol2015vqa} or multimodal reasoning~\citep{yue2024mmmu}.
\method uses fixed instruction templates, which may be suboptimal across product types or tasks. Future work could explore prompt optimization~\citep{pryzant2023automatic} to improve generalization and task alignment.


\section{Ethics Statement}
\dataset is built from publicly available, open-source datasets that are properly licensed for redistribution and research use. To provide user privacy, all user IDs have been fully anonymized, and no identifiable user profile information (e.g., names, addresses) is included.
\bibliography{paper}

\clearpage
\setcounter{table}{0}
\setcounter{figure}{0}
\renewcommand{\thetable}{A\arabic{table}}
\renewcommand{\thefigure}{A\arabic{figure}}

\appendix

\section{Dataset Information}
\label{sec:appendix:data}
\dataset's raw data is sourced from 
\href{https://amazon-reviews-2023.github.io/}{Amazon Review 2023}~\citep{hou2024bridging}, 
\href{https://github.com/amazonqa/amazonqa}{AmazonQA}~\citep{gupta2019amazonqa}, and 
\href{https://github.com/amazon-science/esci-data}{Shopping Query Dataset}~\citep{reddy2022shopping}. The Shopping Query Dataset is under Apache License 2.0, and the others do not specify their licenses. 

We summarize the four core e-commerce capabilities that \dataset covers in Section~\ref{sec:appdix:e-commerce_capabilities}, the eight real-world tasks that measure the core e-commerce capabilities in Section~\ref{sec:appendix:task}, the details of the dataset collection process in Section~\ref{sec:appdix:data_collection}, and the statistics of the dataset in Section~\ref{sec:appendix:statistics}.

\subsection{E-commerce capabilities}
\label{sec:appdix:e-commerce_capabilities}

\begin{figure}[ht]
    \centering
    \includegraphics[trim=20 20 20 20, clip, width=0.75\linewidth]{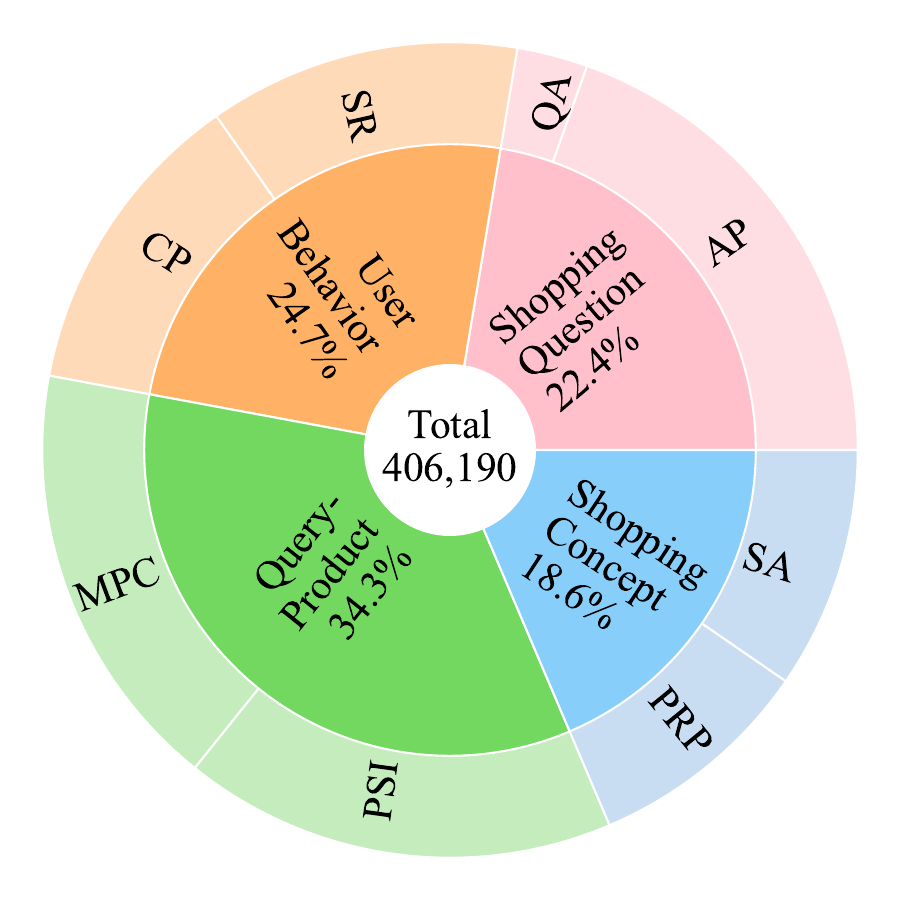}
    \caption{\dataset Capabilities}
    \label{fig:pie_dataset}
\end{figure}

\dataset covers four core e-commerce capabilities: 
\textbf{(1)} shopping question perception, 
%
\textbf{(2)} user behavior alignment, which focuses on modeling and predicting the interactions; 
\textbf{(3)} query-product perception, which captures how relevant products are to customer search queries;
\textbf{(4)}
 and shopping concept understanding, which models the shopping concepts such as product relations and user opinions. 
We summarize the tasks corresponding to each e-commerce capability in
Figure~\ref{fig:pie_dataset}, and introduce these capabilities in detail as follows.

\paragraph*{Shopping Question Perception}Users often ask product-related questions that require both textual and visual understanding (e.g., size, color). We include two tasks---question answerability prediction (\AP) and binary question answering (\QA)~\citep{gupta2019amazonqa}---to evaluate a model’s capability in answering such questions. 

\paragraph*{User Behavior Alignment}Understanding user behavior is essential for optimizing personalized shopping experiences. 
E-commerce platforms utilize user interactions, including browsing and purchase history, to predict future behaviors. 
To assess the models’ ability to align with user behavior, we introduce two tasks: 
click-through prediction (\CP) and sequential recommendation (\SR)~\citep{hou2024bridging}.

\paragraph*{Query-product Perception}Accurately interpreting user searching queries and understanding products' relevance to them is crucial in e-commerce. This capability ensures that users receive the most relevant product suggestions based on their search intent. To evaluate the models’ capability in this area, we include two asks:
multi-class product classification  (\MPC) and product substitute identification (\PSI)~\citep{reddy2022shopping}.

\paragraph*{Shopping Concept Understanding} Online shopping concepts
are important in interpreting products.
Failing to understand these concepts compromises the performance of models on downstream tasks. To examine the models' capability of shopping concept understanding, we introduce two tasks: 
product relation prediction (\PRP) and sentiment analysis (\SA)~\citep{xu2020prp, daza2024sentiment}.

\subsection{Task Definition}
\label{sec:appendix:task}
We provide the task definition below.

\textbf{Answerability prediction (\AP)}~\citep{gupta2019amazonqa}: Predict whether the product-related question is answerable based on the product information.

\textbf{Binary question answering (\QA)}~\cite{gupta2019amazonqa}: given the product-related question, product images, and user reviews, answer the yes-no question or indicate when the question cannot be answered.

\textbf{Click-through rate prediction (\CP)}~\cite{hou2024bridging}: Predict if the user would be interested in the candidate products by analysing the user's purchase history.

\textbf{Sequential recommendation (\SR)}~\cite{kang2018sasrec,hou2024bridging}: Predict the next product that the user would like to buy based on the user’s purchase history.

\textbf{Multi-class product classification  (\MPC)}~\citep{reddy2022shopping}: Given a query and product information, predict relevance between the query and product.

\textbf{Product substitute identification (\PSI)}~\citep{reddy2022shopping}: Predict if the product can serve as a functional substitute for the user’s query.

\textbf{Product relation prediction (\PRP)}~\citep{ni2019justifying, xu2020prp}: Identify the relationship between two product given product information.

\textbf{Sentiment analysis (\SA)}~\citep{wankhade2022sa,daza2024sentiment}: Identify the user's rating that the user would like to give based on the user review information.

\subsection{{Data Collection Details}}
\label{sec:appdix:data_collection}
\dataset is constructed by aggregating raw data from existing, publicly available e-commerce datasets. 
These include structured product metadata, user reviews, question-answer pairs, and behavioral logs from sources 
such as AmazonQA~\citep{gupta2019amazonqa}, Amazon Review 2023~\citep{hou2024bridging}, and the Shopping Queries dataset~\citep{reddy2022shopping}. 
All user identifiers were removed to preserve anonymity.

Importantly, no new human annotation was conducted during the dataset construction process. 
Instead, \dataset follows established practices from instruction-tuning 
and multimodal learning benchmarks~\cite{lee2023platypus, fang2023mol, peng2024ecellm}, 
where task-specific supervision is deterministically derived from structured fields in raw data. 
For instance, in the \PRP task, we extract product relationships such as co-purchase links directly from the ``also\_buy" field. 
Similarly, in \PSI, product similarity and substitutability are derived from structured fields in the predefined schema~\citep{reddy2022shopping}.

Across all 8 benchmark tasks, we apply schema-aligned rules to transform structured inputs into consistent, 
instruction-style prompt-response pairs. An example of raw data used for \PRP is shown below:

\begin{lstlisting}[language=json]
{
  "asin": "0043396828",
  "title": "Books 'Handbook of Astronomical Image Processing' with CD ROM",
  "also_buy": "0999470906"
}
\end{lstlisting}

This schema generates the relation that "Users who buy product 0043396828 may also buy product 0999470906."

To ensure multimodal richness and quality, we aggregate multiple product images, 
including a designated main image and additional views contributed by merchants or users for each item. 
Low-resolution or low-quality images are filtered out. 
The final dataset pairs rich visual input with structured text components 
such as product titles, user queries, reviews, and category information.

The full dataset is then formatted using instruction-style templates, adapted from prior benchmarks such as ECInstruct~\cite{peng2024ecellm} and MMECInstruct~\citep{ling2024caslie}. 
This formatting enables plug-and-play usability for large multimodal models and consistent evaluation across tasks.
Although curated based on publicly available raw sources, the construction of \dataset benchmark is non-trivial, requiring careful design to support multi-image multimodal learning, task diversity, and real-world applicability.

\subsection{Data Statistics}
\label{sec:appendix:statistics}

\begin{figure}[ht]
    \centering
    \includegraphics[width=0.98\linewidth]{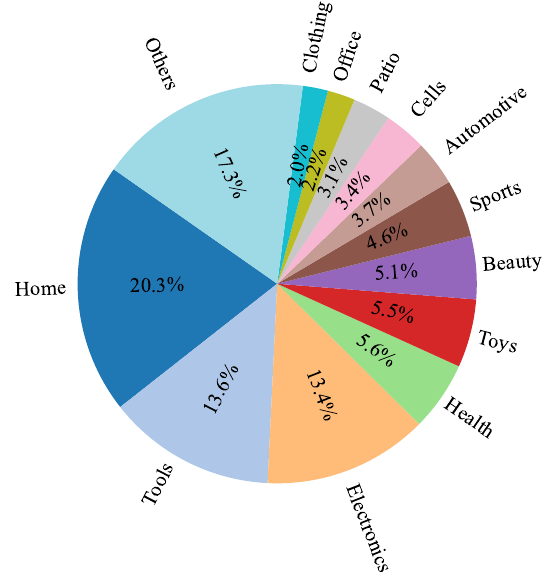}
    \caption{\dataset Category Distribution}
    \label{fig:category}
\end{figure}

The dataset spans products from 34 diverse categories, offering broad coverage such as Home, Electronics, Beauty, Clothing, Automotive, and more. We provide category-level statistics in the Figure~\ref{fig:category}. We focus on presenting the overall dataset scale and task diversity in the current statistics, and we will add these statistics to the revised version.
This distribution supports our goal of creating a representative benchmark for real-world e-commerce tasks, balancing both high-frequency and long-tail product categories.

\subsection{Human Evaluation of \hard}
\label{sec:appdix:vc_set}
To assess the validity of our \hard subset beyond automated model consensus, we conduct a small-scale human annotation study. 
We recruit three independent volunteer annotators with strong educational backgrounds to perform a blind assessment. Due to limited resources, we randomly sample 50 test cases labeled as \hard and ask annotators whether they think the text alone is sufficient to complete the task, and then solve the task. This evaluation protocol is aligned with practices in the prior benchmark study~\citep{chen2024mmstar}.
Table~\ref{tbl:VC_human} summarizes the results.

\begin{table}[!h]
  \centering
  \begin{footnotesize}
  \begin{threeparttable}
      \begin{tabular}{
	@{\hspace{3pt}}l@{\hspace{3pt}}
	@{\hspace{3pt}}r@{\hspace{3pt}}
      }
      \toprule        
       \textbf{Evaluation} & \textbf{Percentage}
        \\
       \midrule
\hard confirmed & 84\% \\
~~lacks attribute info & 38\% \\
~~implicit value & 17\% \\
~~too complex & 33\% \\
      \bottomrule
      \end{tabular}
  \end{threeparttable}
  \end{footnotesize}
  \caption{Human evaluation results of the \hard set.}
  \label{tbl:VC_human}
\end{table}

Results in Table~\ref{tbl:VC_human} show that in 84\% of cases, annotators identify that the textual information is insufficient. Among these, 38\% of the cases lack attribute information (e.g., color, size), while 17\% of the cases contain implicit values that hindered comprehension without visual support. Additionally, 33\% of the cases are deemed too complex for the annotators to resolve.
These observations support the validity of our data-driven annotation method and also highlight that even human annotators may face challenges due to the lack of domain-specific context, underscoring the need for a domain-specific design of multimodal e-commerce modeling.


\section{Method Analysis}
\label{sec:appdix:analysis}
In this section, we discuss the quality of utility annotation, details of evaluation and its results, and real-world considerations.

\subsection{Utility Annotation Analysis}
\label{sec:appdix:vu_annotation}

In the current \method framework, the visual utility assessment (\VUA) and prediction (\VUP) components employ a single e-commerce MLLM to automatically label the utility of input images as helpful, redundant, insufficient, or misleading. This design offers a practical and scalable approach to large-scale annotation, ensuring consistency and reproducibility across millions of multimodal samples.
Nevertheless, because the labeling process depends on the visual reasoning ability of one MLLM, the resulting utility labels may reflect, to some extent, the model’s own interpretive biases. Variations in how different MLLMs process visual cues, such as subtle product attributes or cross-modal relationships, can occasionally lead to noisy or imperfect labels. 
These effects are expected given the evolving nature of multimodal understanding and do not necessarily undermine the overall annotation quality, but they suggest that further improvements are possible.

One potential extension for future work is to explore multimodel or consensus-based annotation strategies, in which multiple MLLMs collaboratively determine image utility, thereby reducing individual model bias and increasing label stability.
Additionally, selectively incorporating human-in-the-loop validation for ambiguous or low-consensus cases could help reinforce semantic grounding where automated models remain uncertain. 
While such extensions would introduce additional cost and complexity, they may further enhance the precision and robustness of visual utility prediction. We leave the systematic investigation of these hybrid annotation approaches to future work.

\subsection{Evaluation Details}
\label{sec:appdix:exp}

While \method is model-agnostic and generalizable in domain-specific design, our focus is deliberately scoped to the multi-image e-commerce setting.
This choice reflects the unique structural characteristics of e-commerce platforms, 
where products are consistently represented by multiple complementary images tied to a single entity (e.g., item variants, usage scenarios). 
In contrast, general multimodal benchmarks~\citep{antol2015vqa, yue2024mmmu} typically involve single-image samples or 
semantically unrelated images across samples, making them not suitable for evaluating \method.
Moreover, \dataset is, to our knowledge, the first large-scale benchmark that 
enables controlled, task-diverse evaluation of visual utility under multi-image settings.
Establishing robust methodology and insight in e-commerce is a critical and underexplored foundation. 
This specificity allows us to precisely measure when and how images help or hinder downstream understanding.

For the compared baselines, recent methods such as Rec-GPT4V~\cite{liu2024rec} and X-Reflect~\cite{lyu2024x} are primarily designed for multimodal recommendation, while CoT prompting~\cite{wei2022chain} targets reasoning in unimodal text contexts. These approaches differ from our setting, which spans diverse multimodal e-commerce tasks such as classification, QA, retrieval, and recommendation. 
Nevertheless, reasoning-oriented techniques like CoT~\cite{wei2022chain} and Reflexion~\cite{shinn2023reflexion} are complementary to our framework and could be integrated into \method to further enhance visual utility prediction and downstream inference, an avenue we leave for future exploration.

\subsection{Full Experimental Results}
\label{sec:appdix:full_result}

We evaluate the performance of fine-tuned \VUP based on the \VUA's assenssment on the general test set of \dataset to investigate how well the predictor learns to assessing visual utiliy.
The result shows that the predictor achieves an accuracy of 0.858 in predicting helpful images,confirming its utility assessment ability and supports its integration into the \method.

Table~\ref{tbl:full_ap},~\ref{tbl:full_qa},~\ref{tbl:full_cp},~\ref{tbl:full_sr},~\ref{tbl:full_mpc},~\ref{tbl:full_psi},~\ref{tbl:full_prp}, and~\ref{tbl:full_sa} present the complete results for \AP, \QA, \CP, \SR, \MPC, \PSI, \PRP, and \SA respectively in both general set and \hard set. Overall, \method models outperform the general LLMs, e-commerce LLMs, general MLLMs, and e-commerce MLLMs.
Note that in all tables, \#Invalid indicates the number of invalid failure cases for which we cannot extract meaningful results from the model output.
We exclude these invalid cases when calculating the metrics, except for accuracy.

\begin{table*}[!h]
  \centering
  \begin{footnotesize}
  \begin{threeparttable}
      \begin{tabular}{
	@{\hspace{3pt}}l@{\hspace{3pt}}
	@{\hspace{3pt}}r@{\hspace{3pt}}
	@{\hspace{3pt}}r@{\hspace{3pt}}
	@{\hspace{3pt}}r@{\hspace{3pt}}
	@{\hspace{3pt}}r@{\hspace{3pt}}
	@{\hspace{3pt}}r@{\hspace{0pt}}
	@{\hspace{10pt}}r@{\hspace{3pt}}
	@{\hspace{3pt}}r@{\hspace{3pt}}
	@{\hspace{3pt}}r@{\hspace{3pt}}
	@{\hspace{3pt}}r@{\hspace{3pt}}
	@{\hspace{3pt}}r@{\hspace{0pt}}
      }
      \toprule        
       \multirow{2}{*}{\textbf{Model}} & \multicolumn{5}{c}{\textbf{\GTS}} & \multicolumn{5}{c}{\textbf{\hard}} \\
       \cmidrule(lr){2-6} \cmidrule(lr){7-11}
       & Acc. & Pre. & Rec. & F1 & \#Invalid & Acc. & Pre. & Rec. & F1 & \#Invalid \\
        \midrule
        \multicolumn{11}{c}{\textit{General and E-commerce LLMs}} \\
        \midrule
Mistral-7B-v0.3 & 0.663 & 0.746 & 0.747 & 0.747 & 3 & 0.458 & 0.851 & 0.498 & 0.629 & 1  \\
Ministral-8B & 0.466 & 0.894 & 0.223 & 0.357 & 0 & 0.055 & 0.278 & 0.018 & 0.033 & 0  \\
Qwen2.5-7B & 0.662 & 0.845 & 0.602 & 0.703 & 0 & 0.211 & 0.731 & 0.225 & 0.344 & 0  \\
Qwen2.5-14B & 0.559 & 0.876 & 0.392 & 0.542 & 0 & 0.059 & 0.404 & 0.049 & 0.087 & 0  \\
Gemma2-9B & 0.641 & 0.823 & 0.588 & 0.686 & 4 & 0.204 & 0.723 & 0.218 & 0.335 & 0  \\
Gemma2-27b & 0.684 & 0.830 & 0.726 & 0.775 & 394 & 0.280 & 0.781 & 0.328 & 0.462 & 107  \\
Phi3.5 & 0.670 & 0.698 & 0.884 & 0.780 & 0 & 0.720 & 0.898 & 0.783 & 0.837 & 0  \\
Phi4 & 0.580 & 0.866 & 0.435 & 0.580 & 0 & 0.070 & 0.461 & 0.067 & 0.117 & 0  \\
eCeLLM-M & 0.812 & 0.848 & 0.875 & 0.861 & 0 & 0.615 & 0.894 & 0.660 & 0.759 & 0  \\
eCeLLM-L & 0.788 & 0.844 & 0.836 & 0.840 & 2 & 0.612 & 0.903 & 0.648 & 0.755 & 1  \\
\midrule
\multicolumn{11}{c}{\textit{General and E-commerce MLLMs}} \\
\midrule
Claude 3.5 & 0.673 & 0.747 & 0.775 & 0.761 & 31 & 0.495 & 0.863 & 0.539 & 0.664 & 7  \\
Phi-vision & 0.532 & 0.782 & 0.410 & 0.538 & 0 & 0.174 & 0.717 & 0.168 & 0.273 & 0  \\
Qwen2-VL-7B & 0.570 & 0.738 & 0.722 & 0.730 & 922 & 0.406 & 0.857 & 0.499 & 0.631 & 189  \\
Llava-interleave & 0.616 & 0.691 & 0.766 & 0.727 & 0 & 0.592 & 0.886 & 0.638 & 0.742 & 0  \\
CASLIE-M & 0.807 & 0.821 & 0.908 & 0.862 & 0 & 0.702 & 0.904 & 0.757 & 0.824 & 0  \\
CASLIE-L & 0.756 & 0.792 & 0.858 & 0.824 & 3 & 0.642 & 0.895 & 0.693 & 0.781 & 1  \\
\midrule
\multicolumn{11}{c}{\textit{Ours}} \\
\midrule
\methodL & 0.717 & 0.704 & 0.991 & 0.823 & 0 & 0.902 & 0.918 & 0.981 & 0.948 & 0  \\
\methodC & 0.819 & 0.827 & 0.922 & 0.872 & 0 & 0.738 & 0.911 & 0.793 & 0.848 & 0  \\
      \bottomrule
      \end{tabular}
  \end{threeparttable}
  \end{footnotesize}
  \caption{Experimental results on the \AP task.}
  \label{tbl:full_ap}
\end{table*}

\begin{table*}[!h]
  \centering
  \begin{footnotesize}
  \begin{threeparttable}
      \begin{tabular}{
	@{\hspace{3pt}}l@{\hspace{3pt}}
	@{\hspace{3pt}}r@{\hspace{3pt}}
	@{\hspace{3pt}}r@{\hspace{3pt}}
	@{\hspace{3pt}}r@{\hspace{3pt}}
	@{\hspace{3pt}}r@{\hspace{3pt}}
	@{\hspace{3pt}}r@{\hspace{0pt}}
	@{\hspace{10pt}}r@{\hspace{3pt}}
	@{\hspace{3pt}}r@{\hspace{3pt}}
	@{\hspace{3pt}}r@{\hspace{3pt}}
	@{\hspace{3pt}}r@{\hspace{3pt}}
	@{\hspace{3pt}}r@{\hspace{0pt}}
      }
      \toprule        
       \multirow{2}{*}{\textbf{Model}} & \multicolumn{5}{c}{\textbf{\GTS}} & \multicolumn{5}{c}{\textbf{\hard}} \\
       \cmidrule(lr){2-6} \cmidrule(lr){7-11}
       & Acc. & Pre. & Rec. & F1 & \#Invalid & Acc. & Pre. & Rec. & F1 & \#Invalid \\
        \midrule
        \multicolumn{11}{c}{\textit{General and E-commerce LLMs}} \\
        \midrule
Mistral-7B-v0.3 & 0.527 & 0.703 & 0.416 & 0.383 & 57 & 0.224 & 0.149 & 0.299 & 0.194 & 21  \\
Ministral-8B & 0.261 & 0.242 & 0.395 & 0.241 & 0 & 0.177 & 0.219 & 0.234 & 0.146 & 0  \\
Qwen2.5-7B & 0.551 & 0.508 & 0.417 & 0.392 & 0 & 0.204 & 0.101 & 0.271 & 0.141 & 0  \\
Qwen2.5-14B & 0.581 & 0.570 & 0.473 & 0.474 & 0 & 0.384 & 0.252 & 0.274 & 0.242 & 0  \\
Gemma2-9B & 0.555 & 0.504 & 0.475 & 0.476 & 17 & 0.293 & 0.215 & 0.225 & 0.217 & 5  \\
Gemma2-27b & 0.570 & 0.513 & 0.496 & 0.495 & 0 & 0.195 & 0.182 & 0.163 & 0.164 & 0  \\
Phi3.5 & 0.430 & 0.463 & 0.450 & 0.415 & 0 & 0.251 & 0.298 & 0.284 & 0.252 & 0  \\
Phi4 & 0.505 & 0.508 & 0.502 & 0.481 & 0 & 0.209 & 0.200 & 0.219 & 0.188 & 0  \\
eCeLLM-M & 0.351 & 0.542 & 0.417 & 0.281 & 0 & 0.200 & 0.124 & 0.267 & 0.161 & 0  \\
eCeLLM-L & 0.234 & 0.513 & 0.363 & 0.211 & 0 & 0.249 & 0.154 & 0.331 & 0.191 & 0  \\
\midrule
\multicolumn{11}{c}{\textit{General and E-commerce MLLMs}} \\
\midrule
Claude 3.5 & 0.540 & 0.478 & 0.435 & 0.423 & 9 & 0.283 & 0.207 & 0.214 & 0.197 & 7  \\
Phi-vision & 0.312 & 0.496 & 0.414 & 0.291 & 0 & 0.192 & 0.208 & 0.254 & 0.169 & 0  \\
Qwen2-VL-7B & 0.483 & 0.492 & 0.503 & 0.472 & 0 & 0.249 & 0.279 & 0.279 & 0.250 & 0  \\
Llava-interleave & 0.324 & 0.532 & 0.393 & 0.265 & 0 & 0.175 & 0.113 & 0.234 & 0.147 & 0  \\
CASLIE-M & 0.438 & 0.519 & 0.429 & 0.346 & 0 & 0.372 & 0.198 & 0.253 & 0.204 & 0 \\
CASLIE-L & 0.329 & 0.206 & 0.410 & 0.274 & 0 & 0.313 & 0.158 & 0.230 & 0.187 & 0 \\
\midrule
\multicolumn{11}{c}{\textit{Ours}} \\
\midrule
\methodL & 0.577 & 0.602 & 0.459 & 0.449 & 0 & 0.384 & 0.308 & 0.278 & 0.244 & 0  \\
\methodC & 0.596 & 0.574 & 0.556 & 0.545 & 0 & 0.305 & 0.347 & 0.322 & 0.295 & 0  \\
      \bottomrule
      \end{tabular}
  \end{threeparttable}
  \end{footnotesize}
  \caption{Experimental results on the \QA task.}
  \label{tbl:full_qa}
\end{table*}

\begin{table*}[!h]
  \centering
  \begin{footnotesize}
  \begin{threeparttable}
      \begin{tabular}{
	@{\hspace{3pt}}l@{\hspace{3pt}}
	@{\hspace{3pt}}r@{\hspace{3pt}}
	@{\hspace{3pt}}r@{\hspace{3pt}}
	@{\hspace{3pt}}r@{\hspace{3pt}}
	@{\hspace{3pt}}r@{\hspace{3pt}}
	@{\hspace{3pt}}r@{\hspace{0pt}}
	@{\hspace{10pt}}r@{\hspace{3pt}}
	@{\hspace{3pt}}r@{\hspace{3pt}}
	@{\hspace{3pt}}r@{\hspace{3pt}}
	@{\hspace{3pt}}r@{\hspace{3pt}}
	@{\hspace{3pt}}r@{\hspace{0pt}}
      }
      \toprule        
       \multirow{2}{*}{\textbf{Model}} & \multicolumn{5}{c}{\textbf{\GTS}} & \multicolumn{5}{c}{\textbf{\hard}} \\
       \cmidrule(lr){2-6} \cmidrule(lr){7-11}
       & Acc. & Pre. & Rec. & F1 & \#Invalid & Acc. & Pre. & Rec. & F1 & \#Invalid \\
        \midrule
        \multicolumn{11}{c}{\textit{General and E-commerce LLMs}} \\
        \midrule
Mistral-7B-v0.3 & 0.570 & 0.550 & 0.761 & 0.639 & 0 & 0.422 & 0.387 & 0.610 & 0.474 & 0  \\
Ministral-8B & 0.532 & 0.800 & 0.084 & 0.152 & 0 & 0.561 & 0.280 & 0.019 & 0.036 & 0  \\
Qwen2.5-7B & 0.511 & 0.862 & 0.025 & 0.049 & 0 & 0.569 & 0.000 & 0.000 & 0.000 & 0  \\
Qwen2.5-14B & 0.592 & 0.750 & 0.273 & 0.401 & 0 & 0.533 & 0.348 & 0.108 & 0.165 & 0  \\
Gemma2-9B & 0.616 & 0.653 & 0.490 & 0.560 & 0 & 0.468 & 0.335 & 0.249 & 0.286 & 0  \\
Gemma2-27b & 0.600 & 0.604 & 0.580 & 0.592 & 2 & 0.435 & 0.346 & 0.367 & 0.356 & 1  \\
Phi3.5 & 0.585 & 0.665 & 0.338 & 0.449 & 0 & 0.487 & 0.278 & 0.127 & 0.175 & 0  \\
Phi4 & 0.587 & 0.640 & 0.394 & 0.488 & 0 & 0.437 & 0.243 & 0.152 & 0.187 & 0  \\
eCeLLM-M & 0.549 & 0.751 & 0.145 & 0.243 & 0 & 0.523 & 0.060 & 0.008 & 0.014 & 0  \\
eCeLLM-L & 0.545 & 0.857 & 0.108 & 0.192 & 0 & 0.568 & 0.400 & 0.027 & 0.051 & 0  \\
\midrule
\multicolumn{11}{c}{\textit{General and E-commerce MLLMs}} \\
\midrule
Claude 3.5 & 0.584 & 0.704 & 0.304 & 0.425 & 19 & 0.519 & 0.345 & 0.140 & 0.200 & 7  \\
Phi-vision & 0.544 & 0.735 & 0.161 & 0.264 & 29 & 0.523 & 0.258 & 0.043 & 0.074 & 14  \\
Qwen2-VL-7B & 0.002 & 0.000 & 0.000 & 0.000 & 1992 & 0.000 & 0.000 & 0.000 & 0.000 & 863  \\
Llava-interleave & 0.573 & 0.556 & 0.712 & 0.625 & 1 & 0.360 & 0.332 & 0.493 & 0.397 & 0  \\
CASLIE-M & 0.582 & 0.731 & 0.581 & 0.647 & 0 & 0.491 & 0.510 & 0.193 & 0.280 & 0 \\
CASLIE-L & 0.552 & 0.608 & 0.566 & 0.586 & 0 & 0.459 & 0.442 & 0.071 & 0.122 & 0 \\
\midrule
\multicolumn{11}{c}{\textit{Ours}} \\
\midrule
\methodL & 0.607 & 0.581 & 0.766 & 0.661 & 0 & 0.449 & 0.404 & 0.612 & 0.487 & 0  \\
\methodC & 0.649 & 0.640 & 0.684 & 0.661 & 0 & 0.529 & 0.458 & 0.561 & 0.504 & 0  \\
      \bottomrule
      \end{tabular}
  \end{threeparttable}
  \end{footnotesize}
  \caption{Experimental results on the \CP task.}
  \label{tbl:full_cp}
\end{table*}

\begin{table*}[!h]
  \centering
  \begin{footnotesize}
  \begin{threeparttable}
      \begin{tabular}{
	@{\hspace{3pt}}l@{\hspace{3pt}}
	@{\hspace{3pt}}r@{\hspace{3pt}}
	@{\hspace{3pt}}r@{\hspace{0pt}}
	@{\hspace{10pt}}r@{\hspace{3pt}}
	@{\hspace{3pt}}r@{\hspace{0pt}}
      }
      \toprule        
       \multirow{2}{*}{\textbf{Model}} & \multicolumn{2}{c}{\textbf{\GTS}} & \multicolumn{2}{c}{\textbf{\hard}} \\
       \cmidrule(lr){2-3} \cmidrule(lr){4-5}
       & Recall@1 & \#Invalid & Recall@1 & \#Invalid \\
        \midrule
        \multicolumn{5}{c}{\textit{General and E-commerce LLMs}} \\
        \midrule
Mistral-7B-v0.3 & 0.002 & 1994 & 0.001 & 1840  \\
Ministral-8B & 0.000 & 2000 & 0.000 & 1844  \\
Qwen2.5-7B & 0.126 & 0 & 0.084 & 0  \\
Qwen2.5-14B & 0.129 & 9 & 0.085 & 9  \\
Gemma2-9B & 0.072 & 193 & 0.058 & 179  \\
Gemma2-27b & 0.100 & 336 & 0.069 & 314  \\
Phi3.5 & 0.073 & 0 & 0.060 & 0  \\
Phi4 & 0.003 & 1978 & 0.001 & 1826  \\
eCeLLM-M & 0.209 & 0 & 0.171 & 0  \\
eCeLLM-L & 0.193 & 0 & 0.159 & 0  \\
\midrule
\multicolumn{5}{c}{\textit{General and E-commerce MLLMs}} \\
\midrule
Claude 3.5 & 0.149 & 123 & 0.105 & 117  \\
Phi-vision & 0.100 & 28 & 0.075 & 28  \\
Qwen2-VL-7B & 0.046 & 1092 & 0.029 & 1010  \\
Llava-interleave & 0.066 & 7 & 0.053 & 7  \\
CASLIE-M & 0.219 & 0 & 0.191 & 0  \\
CASLIE-L & 0.143 & 0 & 0.119 & 0  \\
\midrule
\multicolumn{5}{c}{\textit{Ours}} \\
\midrule
\methodL & 0.173 & 0 & 0.136 & 0  \\
\methodC & 0.231 & 0 & 0.197 & 0  \\
      \bottomrule
      \end{tabular}
  \end{threeparttable}
  \end{footnotesize}
  \caption{Experimental results on the \SR task.}
  \label{tbl:full_sr}
\end{table*}

\begin{table*}[!h]
  \centering
  \begin{footnotesize}
  \begin{threeparttable}
      \begin{tabular}{
	@{\hspace{3pt}}l@{\hspace{3pt}}
	@{\hspace{3pt}}r@{\hspace{3pt}}
	@{\hspace{3pt}}r@{\hspace{3pt}}
	@{\hspace{3pt}}r@{\hspace{3pt}}
	@{\hspace{3pt}}r@{\hspace{3pt}}
	@{\hspace{3pt}}r@{\hspace{0pt}}
	@{\hspace{10pt}}r@{\hspace{3pt}}
	@{\hspace{3pt}}r@{\hspace{3pt}}
	@{\hspace{3pt}}r@{\hspace{3pt}}
	@{\hspace{3pt}}r@{\hspace{3pt}}
	@{\hspace{3pt}}r@{\hspace{0pt}}
      }
      \toprule        
       \multirow{2}{*}{\textbf{Model}} & \multicolumn{5}{c}{\textbf{\GTS}} & \multicolumn{5}{c}{\textbf{\hard}} \\
       \cmidrule(lr){2-6} \cmidrule(lr){7-11}
       & Acc. & Pre. & Rec. & F1 & \#Invalid & Acc. & Pre. & Rec. & F1 & \#Invalid \\
        \midrule
        \multicolumn{11}{c}{\textit{General and E-commerce LLMs}} \\
        \midrule
Mistral-7B-v0.3 & 0.660 & 0.400 & 0.374 & 0.374 & 0 & 0.380 & 0.224 & 0.229 & 0.209 & 0  \\
Ministral-8B & 0.645 & 0.370 & 0.396 & 0.338 & 0 & 0.297 & 0.187 & 0.202 & 0.162 & 0  \\
Qwen2.5-7B & 0.624 & 0.380 & 0.417 & 0.394 & 0 & 0.266 & 0.178 & 0.176 & 0.173 & 0  \\
Qwen2.5-14B & 0.509 & 0.369 & 0.432 & 0.357 & 0 & 0.180 & 0.128 & 0.182 & 0.123 & 0  \\
Gemma2-9B & 0.611 & 0.413 & 0.437 & 0.414 & 1 & 0.219 & 0.121 & 0.143 & 0.123 & 1  \\
Gemma2-27b & 0.657 & 0.504 & 0.434 & 0.437 & 0 & 0.313 & 0.282 & 0.194 & 0.195 & 0  \\
Phi3.5 & 0.607 & 0.410 & 0.364 & 0.372 & 0 & 0.301 & 0.194 & 0.201 & 0.197 & 0  \\
Phi4 & 0.669 & 0.414 & 0.442 & 0.426 & 0 & 0.311 & 0.197 & 0.202 & 0.197 & 0  \\
eCeLLM-M & 0.713 & 0.463 & 0.439 & 0.438 & 0 & 0.443 & 0.282 & 0.273 & 0.258 & 0  \\
eCeLLM-L & 0.661 & 0.413 & 0.408 & 0.409 & 0 & 0.391 & 0.248 & 0.254 & 0.245 & 0  \\
\midrule
\multicolumn{11}{c}{\textit{General and E-commerce MLLMs}} \\
\midrule
Claude 3.5 & 0.685 & 0.491 & 0.390 & 0.408 & 10 & 0.443 & 0.298 & 0.271 & 0.262 & 4  \\
Phi-vision & 0.658 & 0.293 & 0.321 & 0.294 & 0 & 0.418 & 0.122 & 0.241 & 0.161 & 0  \\
Qwen2-VL-7B & 0.658 & 0.403 & 0.374 & 0.316 & 1 & 0.369 & 0.260 & 0.249 & 0.178 & 1  \\
Llava-interleave & 0.645 & 0.349 & 0.276 & 0.248 & 13 & 0.438 & 0.211 & 0.248 & 0.171 & 10  \\
CASLIE-M & 0.723 & 0.606 & 0.486 & 0.525 & 0 & 0.506 & 0.414 & 0.353 & 0.363 & 0  \\
CASLIE-L & 0.674 & 0.493 & 0.411 & 0.421 & 0 & 0.399 & 0.315 & 0.252 & 0.248 & 0  \\
\midrule
\multicolumn{11}{c}{\textit{Ours}} \\
\midrule
\methodL & 0.692 & 0.501 & 0.457 & 0.473 & 0 & 0.438 & 0.335 & 0.292 & 0.299 & 0  \\
\methodC & 0.721 & 0.543 & 0.441 & 0.456 & 0 & 0.481 & 0.338 & 0.289 & 0.277 & 0  \\
      \bottomrule
      \end{tabular}
  \end{threeparttable}
  \end{footnotesize}
  \caption{Experimental results on the \MPC task.}
  \label{tbl:full_mpc}
\end{table*}

\begin{table*}[!h]
  \centering
  \begin{footnotesize}
  \begin{threeparttable}
      \begin{tabular}{
	@{\hspace{3pt}}l@{\hspace{3pt}}
	@{\hspace{3pt}}r@{\hspace{3pt}}
	@{\hspace{3pt}}r@{\hspace{3pt}}
	@{\hspace{3pt}}r@{\hspace{3pt}}
	@{\hspace{3pt}}r@{\hspace{3pt}}
	@{\hspace{3pt}}r@{\hspace{0pt}}
	@{\hspace{10pt}}r@{\hspace{3pt}}
	@{\hspace{3pt}}r@{\hspace{3pt}}
	@{\hspace{3pt}}r@{\hspace{3pt}}
	@{\hspace{3pt}}r@{\hspace{3pt}}
	@{\hspace{3pt}}r@{\hspace{0pt}}
      }
      \toprule        
       \multirow{2}{*}{\textbf{Model}} & \multicolumn{5}{c}{\textbf{\GTS}} & \multicolumn{5}{c}{\textbf{\hard}} \\
       \cmidrule(lr){2-6} \cmidrule(lr){7-11}
       & Acc. & Pre. & Rec. & F1 & \#Invalid & Acc. & Pre. & Rec. & F1 & \#Invalid \\
        \midrule
        \multicolumn{11}{c}{\textit{General and E-commerce LLMs}} \\
        \midrule
Mistral-7B-v0.3 & 0.256 & 0.230 & 0.978 & 0.373 & 0 & 0.052 & 0.053 & 0.853 & 0.099 & 0  \\
Ministral-8B & 0.321 & 0.178 & 0.553 & 0.269 & 0 & 0.028 & 0.001 & 0.010 & 0.001 & 0  \\
Qwen2.5-7B & 0.392 & 0.150 & 0.362 & 0.212 & 0 & 0.104 & 0.000 & 0.000 & 0.000 & 0  \\
Qwen2.5-14B & 0.389 & 0.210 & 0.616 & 0.314 & 0 & 0.144 & 0.009 & 0.117 & 0.016 & 0  \\
Gemma2-9B & 0.303 & 0.203 & 0.713 & 0.316 & 0 & 0.014 & 0.009 & 0.132 & 0.016 & 0  \\
Gemma2-27b & 0.265 & 0.231 & 0.968 & 0.373 & 0 & 0.059 & 0.051 & 0.812 & 0.095 & 0  \\
Phi3.5 & 0.268 & 0.210 & 0.810 & 0.334 & 0 & 0.029 & 0.023 & 0.365 & 0.044 & 0  \\
Phi4 & 0.273 & 0.230 & 0.941 & 0.370 & 0 & 0.044 & 0.043 & 0.685 & 0.081 & 0  \\
eCeLLM-M & 0.798 & 0.745 & 0.249 & 0.373 & 0 & 0.925 & 0.077 & 0.020 & 0.032 & 0  \\
eCeLLM-L & 0.767 & 0.480 & 0.362 & 0.413 & 0 & 0.900 & 0.251 & 0.320 & 0.281 & 0  \\
\midrule
\multicolumn{11}{c}{\textit{General and E-commerce MLLMs}} \\
\midrule
Claude 3.5 & 0.168 & 0.225 & 0.916 & 0.361 & 2924 & 0.032 & 0.046 & 0.721 & 0.087 & 1165  \\
Phi-vision & 0.290 & 0.204 & 0.738 & 0.320 & 0 & 0.030 & 0.016 & 0.244 & 0.030 & 0  \\
Qwen2-VL-7B & 0.308 & 0.185 & 0.601 & 0.282 & 0 & 0.038 & 0.008 & 0.117 & 0.015 & 0  \\
Llava-interleave & 0.258 & 0.212 & 0.843 & 0.339 & 14 & 0.052 & 0.037 & 0.585 & 0.069 & 5  \\
CASLIE-M & 0.802 & 0.597 & 0.379 & 0.464 & 0 & 0.920 & 0.331 & 0.305 & 0.317 & 0  \\
CASLIE-L & 0.750 & 0.408 & 0.334 & 0.367 & 0 & 0.916 & 0.302 & 0.289 & 0.295 & 0  \\
\midrule
\multicolumn{11}{c}{\textit{Ours}} \\
\midrule
\methodL & 0.774 & 0.501 & 0.415 & 0.454 & 0 & 0.907 & 0.292 & 0.371 & 0.327 & 0  \\
\methodC & 0.803 & 0.580 & 0.477 & 0.524 & 0 & 0.924 & 0.393 & 0.447 & 0.418 & 0  \\
      \bottomrule
      \end{tabular}
  \end{threeparttable}
  \end{footnotesize}
  \caption{Experimental results on the \PSI task.}
  \label{tbl:full_psi}
\end{table*}

\begin{table*}[!h]
  \centering
  \begin{footnotesize}
  \begin{threeparttable}
      \begin{tabular}{
	@{\hspace{3pt}}l@{\hspace{3pt}}
	@{\hspace{3pt}}r@{\hspace{3pt}}
	@{\hspace{3pt}}r@{\hspace{3pt}}
	@{\hspace{3pt}}r@{\hspace{3pt}}
	@{\hspace{3pt}}r@{\hspace{3pt}}
	@{\hspace{3pt}}r@{\hspace{0pt}}
	@{\hspace{10pt}}r@{\hspace{3pt}}
	@{\hspace{3pt}}r@{\hspace{3pt}}
	@{\hspace{3pt}}r@{\hspace{3pt}}
	@{\hspace{3pt}}r@{\hspace{3pt}}
	@{\hspace{3pt}}r@{\hspace{0pt}}
      }
      \toprule        
       \multirow{2}{*}{\textbf{Model}} & \multicolumn{5}{c}{\textbf{\GTS}} & \multicolumn{5}{c}{\textbf{\hard}} \\
       \cmidrule(lr){2-6} \cmidrule(lr){7-11}
       & Acc. & Pre. & Rec. & F1 & \#Invalid & Acc. & Pre. & Rec. & F1 & \#Invalid \\
        \midrule
        \multicolumn{11}{c}{\textit{General and E-commerce LLMs}} \\
        \midrule
Mistral-7B-v0.3 & 0.417 & 0.378 & 0.324 & 0.297 & 0 & 0.276 & 0.330 & 0.202 & 0.160 & 0  \\
Ministral-8B & 0.391 & 0.130 & 0.333 & 0.187 & 0 & 0.025 & 0.008 & 0.333 & 0.017 & 0  \\
Qwen2.5-7B & 0.234 & 0.250 & 0.241 & 0.165 & 0 & 0.031 & 0.324 & 0.218 & 0.038 & 0  \\
Qwen2.5-14B & 0.376 & 0.399 & 0.391 & 0.267 & 2 & 0.573 & 0.320 & 0.309 & 0.270 & 0  \\
Gemma2-9B & 0.380 & 0.254 & 0.320 & 0.201 & 2 & 0.033 & 0.233 & 0.230 & 0.023 & 1  \\
Gemma2-27b & 0.386 & 0.280 & 0.330 & 0.192 & 1 & 0.024 & 0.091 & 0.313 & 0.016 & 0  \\
Phi3.5 & 0.388 & 0.130 & 0.331 & 0.187 & 0 & 0.025 & 0.009 & 0.333 & 0.017 & 0  \\
Phi4 & 0.291 & 0.269 & 0.309 & 0.247 & 0 & 0.082 & 0.298 & 0.149 & 0.065 & 0  \\
eCeLLM-M & 0.731 & 0.572 & 0.530 & 0.518 & 0 & 0.843 & 0.379 & 0.468 & 0.377 & 0  \\
eCeLLM-L & 0.654 & 0.544 & 0.550 & 0.545 & 0 & 0.664 & 0.361 & 0.475 & 0.345 & 0  \\
\midrule
\multicolumn{11}{c}{\textit{General and E-commerce MLLMs}} \\
\midrule
Claude 3.5 & 0.258 & 0.418 & 0.461 & 0.265 & 1 & 0.075 & 0.301 & 0.332 & 0.066 & 0  \\
Phi-vision & 0.378 & 0.293 & 0.349 & 0.243 & 0 & 0.036 & 0.302 & 0.309 & 0.036 & 0  \\
Qwen2-VL-7B & 0.427 & 0.315 & 0.304 & 0.295 & 2 & 0.556 & 0.318 & 0.301 & 0.256 & 2  \\
Llava-interleave & 0.390 & 0.164 & 0.335 & 0.193 & 0 & 0.025 & 0.009 & 0.333 & 0.017 & 0  \\
CASLIE-M & 0.744 & 0.612 & 0.556 & 0.553 & 0 & 0.809 & 0.387 & 0.431 & 0.362 & 0  \\
CASLIE-L & 0.674 & 0.450 & 0.500 & 0.470 & 0 & 0.645 & 0.326 & 0.413 & 0.291 & 0  \\
\midrule
\multicolumn{11}{c}{\textit{Ours}} \\
\midrule
\methodL & 0.762 & 0.670 & 0.577 & 0.576 & 0 & 0.812 & 0.430 & 0.534 & 0.393 & 0  \\
\methodC & 0.771 & 0.682 & 0.571 & 0.555 & 0 & 0.807 & 0.393 & 0.488 & 0.362 & 0  \\
      \bottomrule
      \end{tabular}
  \end{threeparttable}
  \end{footnotesize}
  \caption{Experimental results on the \PRP task.}
  \label{tbl:full_prp}
\end{table*}

\begin{table*}[!h]
  \centering
  \begin{footnotesize}
  \begin{threeparttable}
      \begin{tabular}{
	@{\hspace{3pt}}l@{\hspace{3pt}}
	@{\hspace{3pt}}r@{\hspace{3pt}}
	@{\hspace{3pt}}r@{\hspace{3pt}}
	@{\hspace{3pt}}r@{\hspace{3pt}}
	@{\hspace{3pt}}r@{\hspace{3pt}}
	@{\hspace{3pt}}r@{\hspace{0pt}}
	@{\hspace{10pt}}r@{\hspace{3pt}}
	@{\hspace{3pt}}r@{\hspace{3pt}}
	@{\hspace{3pt}}r@{\hspace{3pt}}
	@{\hspace{3pt}}r@{\hspace{3pt}}
	@{\hspace{3pt}}r@{\hspace{0pt}}
      }
      \toprule        
       \multirow{2}{*}{\textbf{Model}} & \multicolumn{5}{c}{\textbf{\GTS}} & \multicolumn{5}{c}{\textbf{\hard}} \\
       \cmidrule(lr){2-6} \cmidrule(lr){7-11}
       & Acc. & Pre. & Rec. & F1 & \#Invalid & Acc. & Pre. & Rec. & F1 & \#Invalid \\
        \midrule
        \multicolumn{11}{c}{\textit{General and E-commerce LLMs}} \\
        \midrule
Mistral-7B-v0.3 & 0.712 & 0.570 & 0.603 & 0.570 & 190 & 0.077 & 0.107 & 0.154 & 0.093 & 57  \\
Ministral-8B & 0.663 & 0.525 & 0.488 & 0.416 & 361 & 0.284 & 0.187 & 0.305 & 0.211 & 93  \\
Qwen2.5-7B & 0.778 & 0.596 & 0.549 & 0.571 & 3 & 0.176 & 0.182 & 0.176 & 0.166 & 0  \\
Qwen2.5-14B & 0.600 & 0.551 & 0.578 & 0.488 & 4 & 0.203 & 0.127 & 0.214 & 0.150 & 2  \\
Gemma2-9B & 0.806 & 0.553 & 0.575 & 0.562 & 1 & 0.097 & 0.112 & 0.103 & 0.097 & 0  \\
Gemma2-27b & 0.814 & 0.656 & 0.577 & 0.618 & 0 & 0.160 & 0.169 & 0.160 & 0.154 & 0  \\
Phi3.5 & 0.012 & 0.419 & 0.317 & 0.226 & 3557 & 0.022 & 0.182 & 0.312 & 0.166 & 504  \\
Phi4 & 0.776 & 0.522 & 0.661 & 0.581 & 116 & 0.112 & 0.141 & 0.145 & 0.132 & 44  \\
eCeLLM-M & 0.794 & 0.652 & 0.677 & 0.662 & 0 & 0.074 & 0.079 & 0.069 & 0.069 & 0  \\
eCeLLM-L & 0.787 & 0.618 & 0.622 & 0.619 & 0 & 0.218 & 0.204 & 0.191 & 0.190 & 0  \\
\midrule
\multicolumn{11}{c}{\textit{General and E-commerce MLLMs}} \\
\midrule
Claude 3.5 & 0.786 & 0.663 & 0.682 & 0.663 & 3 & 0.167 & 0.196 & 0.152 & 0.156 & 1  \\
Phi-vision & 0.710 & 0.570 & 0.534 & 0.492 & 1 & 0.227 & 0.217 & 0.213 & 0.186 & 0  \\
Qwen2-VL-7B & 0.760 & 0.620 & 0.661 & 0.635 & 34 & 0.085 & 0.093 & 0.087 & 0.083 & 3  \\
Llava-interleave & 0.601 & 0.484 & 0.528 & 0.483 & 1 & 0.227 & 0.217 & 0.229 & 0.208 & 0  \\
CASLIE-M & 0.829 & 0.683 & 0.652 & 0.660 & 0 & 0.284 & 0.207 & 0.227 & 0.208 & 0  \\
CASLIE-L & 0.814 & 0.640 & 0.627 & 0.625 & 0 & 0.228 & 0.199 & 0.189 & 0.183 & 0  \\
\midrule
\multicolumn{11}{c}{\textit{Ours}} \\
\midrule
\methodL & 0.820 & 0.668 & 0.659 & 0.662 & 0 & 0.277 & 0.240 & 0.242 & 0.233 & 0  \\
\methodC & 0.830 & 0.693 & 0.678 & 0.681 & 0 & 0.232 & 0.228 & 0.229 & 0.210 & 0  \\
      \bottomrule
      \end{tabular}
  \end{threeparttable}
  \end{footnotesize}
  \caption{Experimental results on the \SA task.}
  \label{tbl:full_sa}
\end{table*}

\subsection{Real-world Considerations}
\label{sec:appdix:real-time}

We further examined the effect of using multiple predicted helpful images instead of a single one in the \method framework to evaluate its real-world impact. The results show an average performance drop of 14.65\% compared with the single-image setting. 
This degradation suggests that simply aggregating multiple helpful images can introduce potential side effects, such as redundant or conflicting visual information, thereby weakening model focus and multimodal alignment. 
In contrast, using a single predicted helpful image enables the model to attend to the most semantically informative visual cue while avoiding interference among partially overlapping visuals. 
These findings reinforce the design choice in \method to selectively incorporate only one image per instance, an approach that proves both more effective and computationally efficient than using multiple images, highlighting the importance of strategic visual utilization over quantity-driven fusion.

\method is designed as a two-stage process at inference: a lightweight predictor (\VUP) first identifies helpful images, followed by a single downstream MLLM inference. In our implementation, the average end-to-end inference time per sample is approximately 840.7 ms. This is acceptable in e-commerce settings, where user-interactive applications such as personalized recommendations or query understanding typically operate within sub-second to multi-second latency budgets. 
Nevertheless, we recognize that there is room for further optimization. Feasible strategies include quantizing or distilling \VUP into lighter predictors, or caching predicted utilities for popular products. These directions are compatible with \method’s modular design and represent promising extensions.

\section{Case Study}

\begin{figure}[ht]
    \centering
    \includegraphics[width=\linewidth]{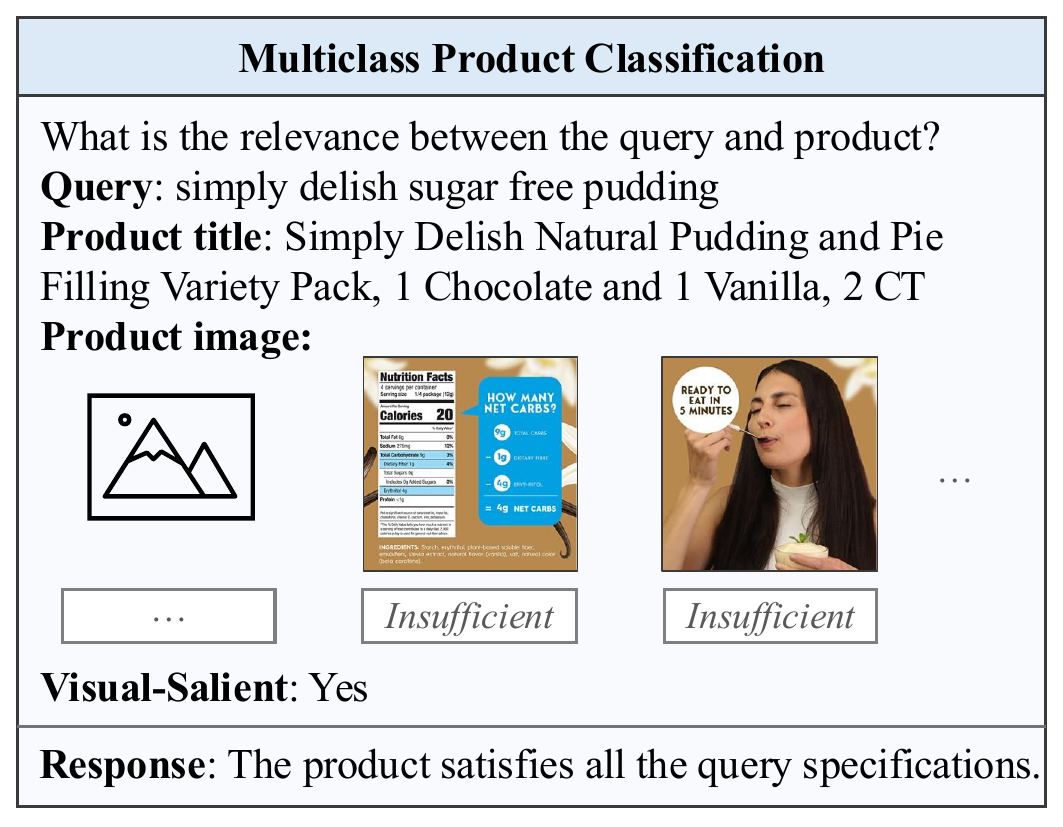}
    \caption{Case study.}
    \label{fig:case_study}
\end{figure}

To have a comprehensive understanding of the visual utility assessment (Section~\ref{sec:pseudo_labeling} and \ref{sec:image_verifier})
used in \method, we present a representative case from the \MPC task shown in Figure~\ref{fig:case_study}. The goal of the \MPC task is to determine the relevance between a product and a user’s query.

The case is accompanied by two images labeled as insufficient for the given example, despite seemingly providing aesthetically relevant or fine-grained information that may assist a human shopper. 
This reflects the distinction between human-perceived informativeness and the practical utility of visual content in MLLMs for a specific task.
For instance, when the user queries “simply delish sugar free pudding”,
%
%
images with heavy characters or tasting scenarios may be visually informative to a scrupulous human. However, due to the current MLLMs’ limited capacity for visual details recognition~\citep{tong2024eyes}, it is hard for MLLMs to effectively utilize such images. 
Though these images may appear useful from a human perspective, 
their detailed visual information cannot be fully utilized by MLLMs and are thus labeled as insufficient.
This design does not imply that the image lacks value in general; rather, it reflects the model’s current inability to extract actionable task-specific information from the visual content.

This case underscores that our visual utility assessment does not denote the general quality or aesthetic relevance of an image. 
Instead, it reflects a model-dependent and task-specific judgment of contribution. 
\method is explicitly designed to reflect such practical model limitations, while highlighting opportunities for improving multimodal comprehension. 
By grounding image filtering decisions in empirical model behavior, our approach enhances both robustness and computational efficiency. 
As vision-language models continue to advance, the definition of these labels may evolve, enabling more effective and comprehensive utilization of complex visual inputs in the future.

\section{Instruction Templates}
\label{sec:appdix:templates}
In this section, we list all instruction templates and few-shot examples used in model inference on the downstream tasks and \hard set selection.

\subsection{Template of \AP}

\texttt{\small{\textbf{Instruction} \\
Analyze the question related to the product and its supporting document, predict if the question is answerable based on the provided document. Only answer yes or no.\\
\textbf{Examples} \\
Input: [question: Will it staple as few as 2 pages?, document: "The stapler is made well and was a fair price. The older model of the same stapler worked a little better. The only bummer is that sometimes when you put the paper in it doesnt click right away. Its heavy duty and can staple lots of paper with no problem but sometimes you need to hold it with two hands. I would buy it again though. Works good 90\% of the time.", "This is exactly like the one we purchased several years ago, we hope we get the same service. It is a good product.", "I love this stapler. I use this stapler on my desk at work.....it's very good at stapling up to 20 pieces of paper.", "I like having an electric stapler at my desk and I want it to work each and every time. This one does that....no jamming. I am very pleased."]. Answer: no.\\
Input: [question: Will this holder work with a Kindle Fire 8.9 Tablet?, document: "I am so glad I got this to use with my iPad mini, and I also use it with my Kindle Fire. It is just what I needed, to sit on the couch with my legs up, as I have a problem knee, and can't sit at the computer for hours a day.", "Love this product HOWEVER there is nothing on the bottom ledge or the backplate to prevent your tablet from sliding off when you move. The non-skid strips on the bottom of the unit don't help when the back plate is raised to accommodate the tablet. Would give 5 stars if my iPad, mini or Kindle wouldn't skid off.", "I bought this as a Christmas gift for my mom. As I write this review, I'm watching her use it across the room in her Kindle Fire. She is 82 and loves to read - and this works great for both hard copy books and electronic books on her Kindle. It's thin, so she can use it in bed or sitting in a chair or on the couch. No more tired arms from holding the Kindle.", "This lap pad for my Kindle is just right for any angle I need. It is very versatile and I got another one for my daughter."]. Answer: yes.
}}

\subsection{Template of \QA}
\texttt{\small{\textbf{Instruction} \\
Given the question related to the product and its supporting document, answer the yes-no question. Please indicate when the question cannot be answered. Only output yes, no, or cannot answer.\\
\textbf{Examples} \\
Input: [question: does it giggle, document: "She is in love with Elmo and immediately started giggling with Elmo's giggle. It's  a very cute toy. Very soft, too.", "Warning : very sensative so it will giggle with even the smallest touch- other than that my 2 yr old niece loves this! A bit smaller than I imagined it to be.", "if your child loves balls and elmo, this is the toy to get, giggles for everyone. Great play for all.", "Perfect size and she loves it. If there is anything wrong is that you cant turn it off. But that was not an issue", "My son loved this ball so much...he loves Elmo's voice. He will lay on his back and just shake it. Super light and easy for my 10 month old to pick up and throw!", "My mother also loved this gift as well! She didn't expect the ball to laugh! :D Great ass for her collection"]. Answer: yes.\\
Input: [question: So building regular exia is possible in this kit?, document: "I hadn't seen much of Gundam Exia, and my favorite Gundam was the 00 Raiser. Now that I am collecting model kits, I'll say, that Exia is my favorite Gunda.", "Sadly like most models I have bought this guy comes with a beam that you need to color, and this one is actually useful because he has this piece where he can stick the beams on his foot armor. Overall this is my favorite exia model and i would recommend for any Gundam fan 00 or not b/c this guy ROOCCCKSS!", "This thing is very cool he has is very colorful when you put on the stickers because the yellow really brings out his beauty. It comes with a stand that you will HAVE to use if you want to pose this dude with the leg armor on. ", "this is a very good model with the gundam. I like it very much. I think it is worth at these money.", "Some of the parts are fairly lose, overall this figure is a good buy and i would recommend buying it."]. Answer: cannot answer.
}}

\subsection{Template of \CP}
\texttt{\small{\textbf{Instruction} \\
Analysis the user's purchase history, predict if the user would like to buy the candidate product. Only output yes or no.\\
\textbf{Examples} \\
Input: [purchase hisroty: "Product 1: X3 Industrial Blue Nitrile Gloves, Box of 200, 3 Mil, Size Large, Latex Free, Powder Free, Textured, Disposable, Non-Sterile, Food Safe, X3D46100BX,X3D46100-BX. Product 2: Brita Water Filter Replacements for Sink, Faucet Mount Water Filtration System for Tap Water, Reduces 99\% of Lead, Chrome, 2 Count. Product 3: Brita Water Filter for Sink, Complete Faucet Mount Water Filtration System for Tap Water, Reduces 99\% of Lead, White. Product 4: DEWALT 20V MAX* XR Rotary Hammer Drill, D-Handle, 1-Inch, Tool Only (DCH133B).", candidate product: "Prescott Plastics 0.5 Inch Square Plastic Plug Insert (20 Pack), Black End Cap for Metal Tubing, Fence, Glide Insert for Pipe Post, Chairs and Furniture."]. Answer: yes.\\
Input: [purchase hisroty: "Product 1: Coleman Cable 23538805 Cord Set, 16/3 80' SJTW Green. Product 2: Lutron Maestro Motion Sensor Switch, 2 Amp, Single Pole, MS-OPS2-WH, White. Product 3: Lutron Maestro LED+ Motion Sensor Dimmer Switch | No Neutral Required | MSCL-OP153M-BI | Biscuit. Product 4: KOHLER K-5282-NA Strive 35-1/2 x 20-1/4 x 9-5/16 Undermount Double-Bowl Kitchen Sink with Basin -Rack, X-Large/Medium, Stainless Steel. Product 5: Brita Standard Water Filter Replacements for Pitchers and Dispensers, Lasts 2 Months, Reduces Chlorine Taste and Odor, 4 Count.", candidate product: "Aigostar LED USB Rechargeable Book Light for Reading in Bed, Clip on Reading Book Light with Gooseneck,Desk Table Lamp with 3 Brightness Dimmable,Eye Care Portable Bed Bedside Clamp Light White."]. Answer: no.
}}

\subsection{Template of \SR}
\texttt{\small{\textbf{Instruction} \\
Estimate the user's intent based on the user's purchase history, and predict the next product that the user is most likely to purchase from the given options. Only answer from options.\\
\textbf{Examples} \\
Input: ["Product 1: DreamSpa 3-way Shower Combo PLUS Instant-Mount Drill-Free Slide Bar - Enjoy Overhead \& Handheld Shower Head with Height\/Angle Adjustable Bracket and Stainless Steel Hose for Ultimate Convenience! Product 2: Ivation Motion-Activated Outdoor Solar-Powered Floodlight with 53 Bright LEDs. Product 3: Moen SMA1005CH Home Care Securemount Anchor, 1 Anchor, Chrome. Product 4: Globe Electric 6-Light Adjustable S-Shape Track Lighting, Black Color, Bulbs Included, Track Lighting Kit, Ceiling Light Fixture."]. Options: [A: SE Deluxe Butane Power Torch with Built-In Ignition System - MT3001.; B: DEWALT, DCG412B, 20V MAX Cut-Off Tool ONLY includes Unit Instruction Guide.; C: Spectrum Diversified OTC\/OTD Paper Towel Holder, 5H x 11-3\/4W x 1-5\/8D, Brushed Nickel.; D: Design House 516732 Monterey 1 Light Wall Light, Oil Rubbed Bronze.; E: Dead On DO-FR Framers Rig, 1 Size Fits All.]. Answer: C.\\
Input: ["Product 1: SadoTech Wireless Doorbells for Home, Apartments, Businesses, Classrooms, etc. - 1 Door Bell Ringer \& 2 Plug-In Chime Receiver, Battery Operated, Easy-to-Use, Wireless Doorbell w\/LED Flash, White. Product 2: BOSCH GLM 15 Compact Laser Measure, 50-Feet. Product 3: ZEEFO Retro LED Night Light Wireless PIR Motion Sensor Light,Activated Step lighting Lamps,Indoor\/Outdoor Battery-Operated Light-Sensitive Portable Moving Table Lamp for Kids Room,Hallway. Product 4: Big Floors GarageTrac Diamond, Durable Copolymer Interlocking Modular Non-Slip Garage Flooring Tile, Red."]. Options: [A: GE Ultrabrite LED Night Light, Dimmable, 100 Lumens, Plug-in, Dusk to Dawn Sensor, UL-Listed, Ideal for Bedroom, Bathroom, Nursery, Kitchen, Hallway, 46789, White, 2 Count.; B: Himalayan Glow SPT-2 Original Replacement Electric Cord with (ETL Certified) Dimmer Switch.; C: Heavy Duty Professional 2-13mm 1\/2-inch SDS-Plus Keyless Adapter with Drill Chuck.; D: goldenwarm Pocket Door Lock, Matte Black Contemporary Privacy Square Pocket Door Hardware, Black Sliding Pocket Door Lock(1 Pack).]. Answer: A.
}}

\subsection{Template of \MPC}
\texttt{\small{\textbf{Instruction} \\
Predict the relevance between the query and product by analyzing the user's query and product title. Here are the options: ['A: The product is relevant to the query, and satisfies all the query specifications.', 'B: The product is somewhat relevant. It fails to fulfill some aspects of the query but the product can be used as a functional substitute.', 'C: The product does not fulfill the query, but could be used in combination with a product exactly matching the query.', 'D: The product is irrelevant to the query.']. Output the option that best describes the relevance. Only answer from options.\\
\textbf{Examples} \\
Input: [query: st ives pink lemon and mandarin orange scrub, product title: St Ives Body Wash 24 Ounce Radiant (Lemon\/Mandarin) (709ml) (3 Pack)]. Answer: C.\\
Input: [query': baby swings for infants clearance', product title: Fisher-Price Infant-to-Toddler Rocker Floral Confetti, stationary baby seat and rocking chair with toys]. Answer: B.
}}

\subsection{Template of \PSI}
\texttt{\small{\textbf{Instruction} \\
Given a user's query and product information, identify if the product is somewhat relevant to the query but fails to fulfill some aspects of the query but the product can be used as a functional substitute. Only answer yes or no.\\
\textbf{Examples} \\
Input: [query: patek philips watch women, product title: Bruno Magli Women's Valentina 1064 Swiss Quartz Italian Leather Strap Watch (White)]. Answer: yes.\\
Input: [query: red dot sight sightmark, product title: Sightmark Wolverine FSR LQD Red Dot Sight,SM26020-LQD]. Answer: no.
}}

\subsection{Template of \PRP}
\texttt{\small{\textbf{Instruction} \\
Given the information of two products, predict the relation of the two products. Here are the options: ['A: Users who buy product 1 may also buy product 2.', 'B: Users who view product 1 may also view product 2.', 'C: The product 1 is similar with the product 2.']. Only answer from options.\\
\textbf{Examples} \\
Input: [product 1 title: DEWALT DC821KA 18-Volt 1/2-Inch Compact Impact Wrench, product 2 title: DEWALT 20V Max XR Impact Wrench Kit, Brushless, High Torque, Detent Pin Anvil, 1/2-Inch, Cordless (DCF899M1)]. Answer: B.\\
Input: [product 1 title: Wilton B.A.S.H 12" Sledge Hammer, 2.5 Lb (20212), product 2 title: Crescent 14 Piece 12 Point SAE Combination Wrench Set with Tool Roll - CCWS4]. Answer: A.
}}

\subsection{Template of \SA}
\texttt{\small{\textbf{Instruction} \\
Given the user's review information, identify the user's sentiment. Answer from the options: ['A: very positive', 'B: positive', 'C: neutral', 'D: negative', 'E: very negative'].\\
\textbf{Examples} \\
Input: [review: Awesome. But cheap hardware It's a nice product and works well but they came with cheap screws which broke midway screwing it in the wall. Which now I cant take the broken screw out.]. Answer: B.\\
Input: [review: Received them in a good packaging, no damages I purchased a 12 pack of these lights. Product was as described. Received them in a good packaging, no damages.  Customer service from the seller was great.  They answered all my questions over the phone prior to placing the order. Very easy to install; fits perfectly into the housing (retrofit).  Works great with the LED dimmer.  There is no noise from any of the bulbs.]. Answer: A.
}}

\section{Model Size and Budget}
\label{sec:appendix:budget}
The model size and budget are reported in Table~\ref{tbl:budget}.
\begin{table}[!h]
  \centering

  \footnotesize
  \begin{threeparttable}
      \begin{tabular}{
        @{\hspace{2pt}}l@{\hspace{6pt}}
        @{\hspace{6pt}}r@{\hspace{2pt}}
        @{\hspace{3pt}}r@{\hspace{2pt}}
      }
      \toprule
      \textbf{Model} & \textbf{GPU Memory} & \textbf{Training Time} \\
      \midrule
      \methodL& 25B & 11.5h \\
      \methodC & 15B & 4.0h \\
      \bottomrule
      \end{tabular}
  \end{threeparttable}
    \caption{Model budget and size.}
      \label{tbl:budget}
    \end{table}

\end{document}